\documentclass[10pt]{article}
\usepackage[top=0.85in,left=0.75in,footskip=0.75in,marginparwidth=2in]{geometry}

\usepackage[utf8]{inputenc}

\usepackage{biblatex}
\usepackage{algorithm}
\usepackage[noend]{algpseudocode}
\usepackage{multirow}
\usepackage{graphicx}
\usepackage{adjustbox}
\usepackage{subfig}

\newcommand{\mname}{{\em FERMULEX}}
\newcommand{\rwA}{{\em Struct}}
\newcommand{\rwB}{{\em MaxSim}}
\newcommand{\rwC}{{\em ExtCand}}
\newcommand{\rwD}{{\em CDI-MaxSim}}

\usepackage{nameref,hyperref}

\usepackage[right]{lineno}

\usepackage{microtype}
\DisableLigatures[f]{encoding = *, family = * }

\usepackage{setspace} 

\usepackage{changepage}

\makeatletter
\renewcommand{\@biblabel}[1]{\quad#1.}
\makeatother

\usepackage{lastpage,fancyhdr,graphicx}
\usepackage{epstopdf}
\fancyheadoffset[L]{2.25in}
\fancyfootoffset[L]{2.25in}

\usepackage{color}

\definecolor{Gray}{gray}{.25}

\usepackage{sidecap}

\usepackage{wrapfig}
\usepackage[pscoord]{eso-pic}
\usepackage[fulladjust]{marginnote}
\reversemarginpar

\begin{document}
\vspace*{0.35in}

\begin{flushleft}
{\Huge
\textbf\newline{Feature-rich multiplex lexical networks reveal mental strategies of early language learning}
}
\newline
\\
Salvatore Citraro \textsuperscript{1,2},
Michael S. Vitevitch \textsuperscript{3},
Massimo Stella \textsuperscript{4}\textsuperscript{+},
Giulio Rossetti \textsuperscript{2}\textsuperscript{+}

\bigskip
\bf{1} Department of Computer Science, University of Pisa
  Largo Bruno Pontecorvo, 3, Pisa
\\
\bf{2} KDD-Lab, ISTI (CNR) G. Moruzzi, 1, Pisa
\\
\bf{3} Department of Psychology, University of Kansas, USA
\\
\bf{4} CogNosco Lab, Department of Computer Science, University of Exeter, UK
\\
\bigskip
Corresponding author: m.stella AT exeter.ac.uk
\\
\bigskip

\footnotesize \bf{+} These authors contributed equally.


\end{flushleft}

\section*{Abstract}
Knowledge in the human mind exhibits a dualistic vector/network nature. Modelling words as vectors is key to natural language processing, whereas networks of word associations can map the nature of semantic memory. We reconcile these paradigms - fragmented across linguistics, psychology and computer science - by introducing FEature-Rich MUltiplex LEXical (FERMULEX) networks. This novel framework merges structural similarities in networks and vector features of words, which can be combined or explored independently. Similarities model heterogenous word associations across semantic/syntactic/phonological aspects of knowledge. Words are enriched with multi-dimensional feature embeddings including frequency, age of acquisition, length and polysemy. These aspects enable unprecedented explorations of cognitive knowledge. Through CHILDES data, we use FERMULEX networks to model normative language acquisition by 1000 toddlers between 18 and 30 months. Similarities and embeddings capture word homophily via conformity, which measures assortative mixing via distance and features. Conformity unearths a language kernel of frequent/polysemous/short nouns and verbs key for basic sentence production, supporting recent evidence of children's syntactic constructs emerging at 30 months. This kernel is invisible to network core-detection and feature-only clustering: It emerges from the dual vector/network nature of words. Our quantitative analysis reveals two key strategies in early word learning. Modelling word acquisition as random walks on FERMULEX topology, we highlight non-uniform filling of communicative developmental inventories (CDIs). Conformity-based walkers lead to accurate (75\%), precise (55\%) and partially well-recalled (34\%) predictions of early word learning in CDIs, providing quantitative support to previous empirical findings and developmental theories. 

\doublespacing

\section*{Introduction}

The mental lexicon is the part of memory that stores information about a word's meanings, syntactic features, pronunciation and more \cite{vitevitch2019can,zock2015words,aitchison2012words}.  Although often described as being like a mental dictionary  \cite{elman2004alternative,hills2021networks,vitevitch2019can}, the mental lexicon is not static, and is instead  a complex system, whose structure influences language processing and has been investigated across fields like psychology \cite{vitevitch2019can}, linguistics \cite{aitchison2012words,castro2020contributions}, computer science and artificial intelligence \cite{stella2019modelling,beckage2019network,beckage2016language}. Decades of multidisciplinary research have gathered evidence that words in the mental lexicon have a dual representation \cite{hills2021networks}, analogous to the particle/wave duality of light in physics \cite{beck2008mind}. Psycholinguistics and distributional semantics posit that words in the lexicon possess both a networked organisation \cite{collins1975spreading,quillian1967word,vitevitch2008can} and a vector-space nature \cite{boleda2020distributional,gunther2019vector,landauer1997solution,lenci2018distributional}. On the one hand, networks capture conceptual relationships (as links) between words (as nodes). On the other hand, vector-spaces identify alignment and distances between vectors, whose components represent word features. The network aspects of the mental lexicon started with seminal work by Quillian \cite{quillian1967word} and by Collins and Loftus \cite{collins1975spreading}. These works showed how in a network of words linked through semantic associations, e.g. possessing a common attribute or overlapping in meaning, the length of the shortest path separating concepts was predictive of retrieval times from semantic memory and sentence understanding \cite{collins1975spreading,quillian1967word}. The advent of network science has revived interest in this approach \cite{castro2020contributions}, with several recent works examining how the structure of semantic networks \cite{steyvers2005large,cancho2001small,siew2019cognitive,kumar2021critical,kenett2019can}, phonological networks \cite{vitevitch2008can,vitevitch2014using}, and their multiplex combination \cite{stella2017multiplex,stella2018multiplex,levy2021unveiling} influence language acquisition and processing.

In parallel, distributional semantics postulates that semantic memory possesses a vector space structure \cite{gunther2019vector,boleda2020distributional,pennington2014glove}, where concepts are vectors whose components express either interpretable features \cite{de2008word} (e.g. possessing a semantic feature, being in a category or being acquired at a certain age) or latent aspects of language \cite{jackson2021text,pennington2014glove,lund1996producing,landauer1997solution} (e.g. overlap in meaning due to word co-occurrence in the same context). Although latent aspects of language limit the understanding of cognitive processing, models like Latent Semantic Analysis \cite{landauer1997solution} and the Hyperspace Analogue to Language \cite{lund1996producing} were used extensively in cognitive inquiries of information processing, mainly due to their ability to extract semantic features without human intervention. More recently, transformer neural networks like BERT enabled vector representations for words depending on their context \cite{boleda2020distributional}. This enhancement revolutionised the field of natural language processing and predicted successfully semantic tasks like entity recognition or word meaning disambiguation \cite{boleda2020distributional,jackson2021text}. Although powerful predictors, these approaches provide relatively little access to the organisation of words in the human mind and can thus benefit from network models and interpretable distributional semantics \cite{jackson2021text}. Reconciling the non-latent, interpretable vector/network duality of words in the mental lexicon is the focus of this work.

We introduce FEature-Rich MUltiplex LEXical - \mname\ - networks, which combines the vector-based and multiplex network aspects of words and their associations in the mental lexicon. Rather than merely building networks out of similarities between vectors of features \cite{comin2020complex}, we view structure and feature similarities as two independent building blocks, whose contribution to represent words in the mind can be explored in parallel. Hence in \mname\ networks, network structure remains and can be explored even when word similarities are switched off, and vice versa. This possibility does not exist in networks built from vector similarities (cf. \cite{veremyev2019graph}). We achieve this advancement by using the recent measure of conformity \cite{rossetti2021conformity}, an enhancement of assortative mixing estimation through non-adjacent nodes.

We show that the dual network/vector representation of words is crucial for understanding key aspects of the mental lexicon that would go undetected by considering features - or networks - only. Using normative word learning norms \cite{macwhinney2000childes} and phonological/semantic/syntactic \cite{stella2017multiplex} data in 1000 English toddlers, \mname\ networks reveal a language kernel progressively built in the mental lexicon of toddlers and \textit{undetectable} by either network core detection \cite{holme2005core} or clustering in vector spaces \cite{whelan2015understanding}. This mental kernel contains general yet simple nouns and verbs that can build diverse sentences, with crucial relevance to children's communication \cite{hadley2018sentence}. The identification of this kernel via \mname\ provides quantitative evidence and modelling insights as to how can young children produce early sentences, as recently observed  \cite{hadley2018sentence}.

Modelling word acquisition as increasingly biased random walkers over the network/vectorial \mname\ representation leads to more insights. We adopted this approach inspired by past work using random walkers over cognitive networks for investigating the mental lexicon \cite{griffiths2007google}. We find that predicting word learning in the language kernel crucially depends on: (i) network/vectorial conformity \cite{rossetti2021conformity} and (ii) the filling of communicative developmental inventories (CDIs) \cite{fenson2007macarthur}, i.e. lists of words sharing a semantic category and commonly used for measuring early cognitive development. We find that CDIs display a rich filling dynamic in word learning, which can be predicted by \mname\ with accuracy, precision and recall up to 75\%, 55\% and 34\%, respectively. These values are statistically significant with respect to a baseline random learner. Without conformity and CDI filling levels, in fact, predictions of word learning in the language kernel are equivalent to random guessing. Since the language kernel stores words crucial for producing early sentences, our results indicate that the documented ability for young toddlers to communicate via early sentences around month 30 \cite{hadley2018sentence} crucially depends on network, vector, and categorical aspects of the mental lexicon. Our approach with \mname\ can encompass them all and thus represents a powerful tool for future cognitive research of various aspects of language.

\begin{figure*}[h!]
\centering
  {\includegraphics[scale=0.28]{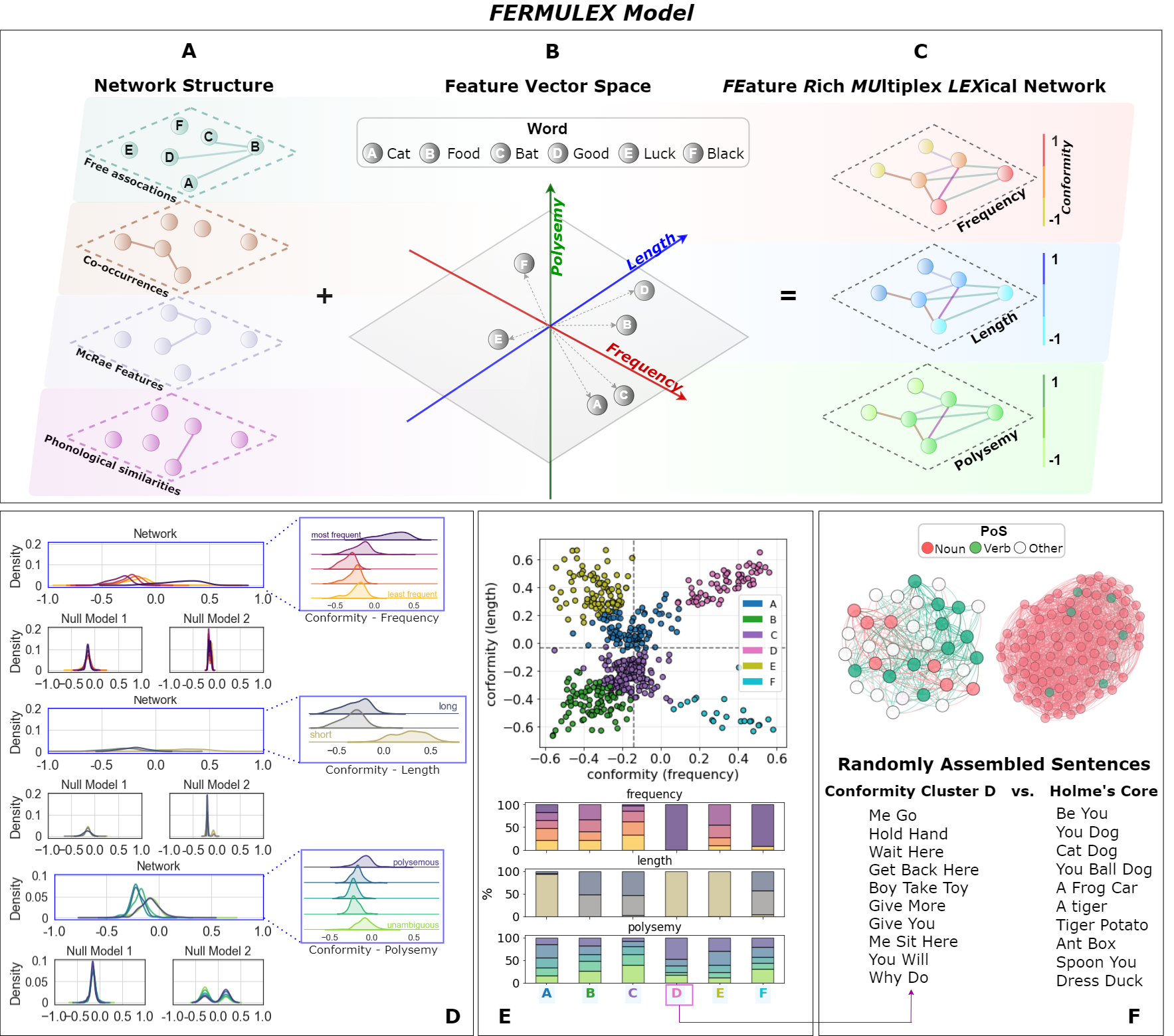}}
    \caption{(A-C): Combining multiplex topology (A) and vector spaces (B) results in \mname\ network (C); (D): kernel density estimates (KDEs) and ridgeline plots highlight conformity distribution for the frequency, length, and polysemy features in toddlers' mental lexicon and the randomised variants; (E): Above -- two-dimensional scatter plot of conformity vector space, where each point is colored according to the cluster the point belongs to (K-means algorithm); Below --
    distribution of word features within each cluster, where a kernel language emerges, i.e. the cluster labeled as \textit{D}; (F): content characterisation of the kernel compared to a competitor from a k-core decomposition.}
    \label{fig:fig_1}
\end{figure*}

\section{Results}

\noindent {\bf \mname\ characterisation}. A combination of a multiplex network structure (Fig. \ref{fig:fig_1}, A) and a vector space of interpretable features (Fig. \ref{fig:fig_1}, B) results in a \mname\ network (Fig. \ref{fig:fig_1}, C).
Conformity \cite{rossetti2021conformity} assesses structure-feature relationships on the aggregated topology.
For each node and with respect to each feature, conformity quantifies the node assortative mixing, by extending this estimation to the non-adjacent but still reachable neighbors of a node.
Studying conformity distributions, we can capture heterogeneous patterns between nodes.

Fig. \ref{fig:fig_1}, D sums up these patterns on the real data representing toddlers' mental lexicon (see Methods for details on network layers and vectors of word features). 
Conformity with respect to frequency highlights an assortative mixing pattern but limited only to highly frequent words, i.e. only words occurring many times in child-directed speech tend to connect with each other in children's \mname\ network. This effect is absent in lower-frequency words and it was not detected in single-layer semantic networks of adults \cite{van2015examining}. Conformity of word length highlights an assortative mixing pattern of very short words only. These two effects are expected to be related as shorter words tend to be more frequent in language \cite{stella2018multiplex}. 

Interestingly, conformity quantifies that polysemous words are likely to connect to each other to a smaller extent than most frequent and shortest words. This indicates an organisation of concepts where unambiguous/less polysemous words are linked to ambiguous/more polysemous words. This heterogeneous mixing by polysemy could be beneficial in providing context and differentiating among possible meanings of a polysemous word, as suggested by previous studies \cite{casas2018polysemy,van2015examining}. If all ambiguous words were grouped together, sense disambiguation could not rely on links including less polysemous/unambiguous words and this homogeneity would ultimately violate the frequency-meaning law \cite{ferrer2018origins}.

The above assortative mixing patterns are not a consequence of feature/distance distributions, because reshuffling node labels (\textit{Null Model 1}) and rewiring links (\textit{Null Model 2}) disrupt the heterogeneous mixing behaviour among classes (see Methods and SI). Hence, the above patterns indicate the presence of a core-periphery organisation in the dualistic multiplex/feature-rich structure of the mental lexicon: A set of highly frequent/shorter/polysemous words linked with each other creates a network core highlighted by conformity and invisible to previous inquiries \cite{stella2017multiplex,stella2019modelling}. This preliminary evidence calls for further analysis of the core.

Fig. \ref{fig:fig_1}, E introduces an analysis of the core performed on: (i) dualistic network/vector and (ii) individual aspects of words in the mental lexicon of toddlers (see Methods and SI). We aim to find a language core that contains groups of words sharing similar structure-feature relationships.
Among the six optimal clusters found (see Methods and SI), groups A and B  (blue and gold) contain mostly short words. Cluster F (cyan) contains highly frequent words. Cluster D contains short, highly frequent and a relevant portion of polysemous words. Sets of clustered words with such features are known as language kernels in cognitive network science \cite{cancho2001small,stella2018multiplex,steyvers2005large}. Language kernels facilitate communication through a small set of simple words suitable for expressing ideas in multiple contexts \cite{cancho2001small}. The conformity core (cluster D) satisfies this definition. In fact, 13\% of the core is made of nouns, 33\% of verbs and the other 54\% include adjectives, adverbs and pronouns, which make it more likely to assemble syntactically well-formed sentences by random sampling compared to other word clusters (cf. Fig. \ref{fig:fig_1} F). Identifying a network core via k-core decomposition \cite{holme2005core} shows almost no meaning organisation and more expressions that are syntactically unrelated (see two random samples in Fig. \ref{fig:fig_1}, F). Analogously, K-Modes \cite{huang1997clustering} attribute-only clusters are unable to form syntactically coherent bigrams. See SI for an analysis centered on computing the internal syntactic coherence of the cores. These comparisons provide unprecedented evidence showing a syntactically advantageous organisation of words in early children's lexicon. This phenomenon goes undetected unless both the network and vector nature of words in the mind is considered.

\begin{figure*}[h!]
\centering
  {\includegraphics[scale=0.3]{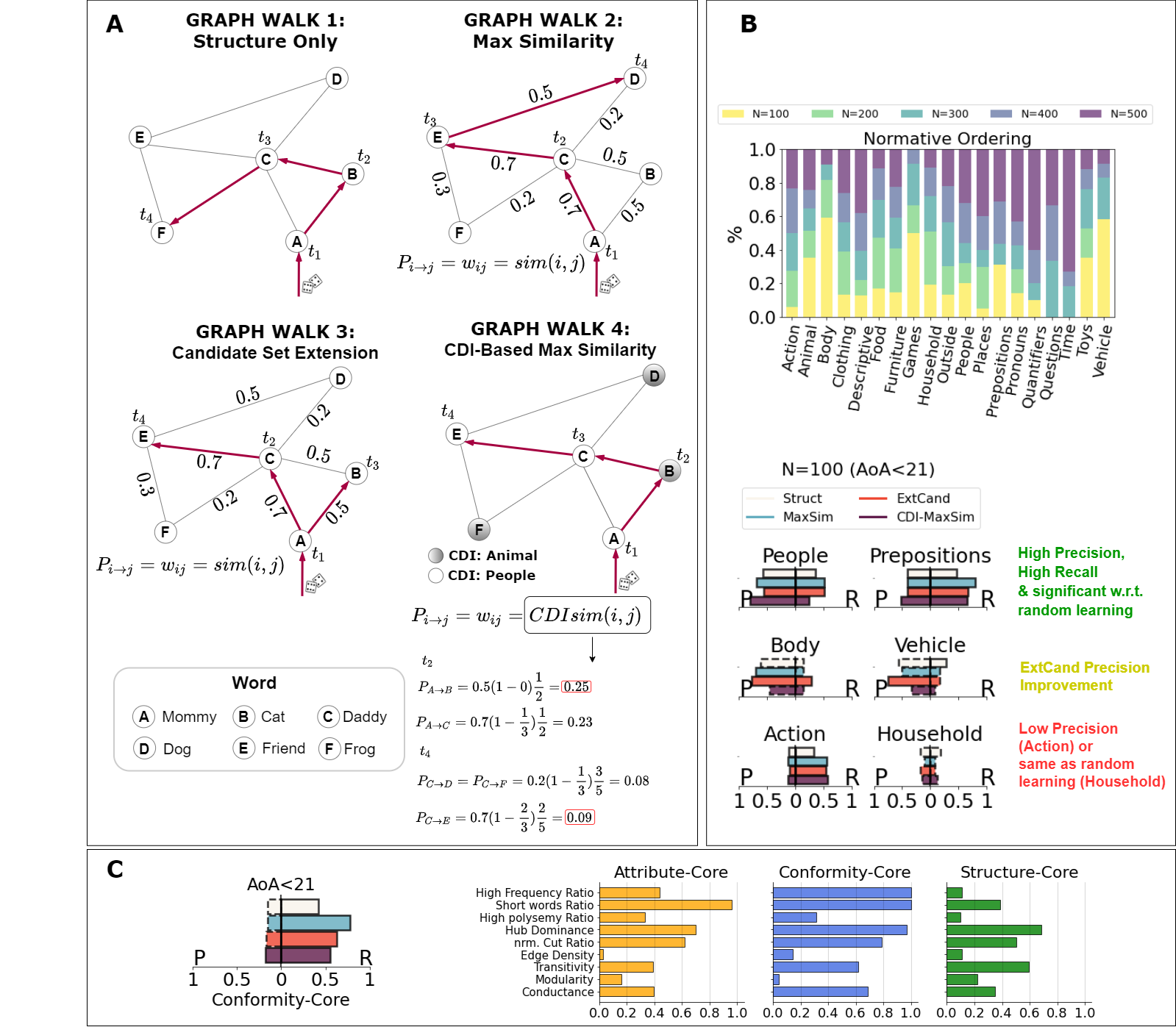}}
    \caption{(A): Random walks combining progressively structure and vector information (Graph Walk 1-3) and CDIs integration (Graph Walk 4); (B): Above -- CDIs filling in CHILDES normative learning; bars show that CDIs are not uniformly filled over time, e.g. more than half of \textit{Body} and \textit{Vehicle} categories are learned during early stage acquisition, whereas \textit{Questions} and \textit{Time} emerge later; Below -- precision-recall evaluation over selected CDIs; solid bars identify statistically significant scores compared to a random learning baseline; (C): Left -- precision-recall evaluation with respect to early acquisition of kernel words; Right -- kernel characterisations using several quality measures.}
    \label{fig:fig_2}
\end{figure*}

\noindent {\bf Topology and cognitive relevance of the conformity core \mname}. We further compare the conformity core with the k-core decomposition \cite{holme2005core} (where similarities are switched off) and with the most relevant K-Modes cluster (where network structure is switched off). Interestingly, the conformity core appears to be a synthesis of the other two potential language kernels. 
Fig. \ref{fig:fig_2}, C characterises the three cores with several qualitative functions assessing intra-cluster connectivity and inter-cluster distinctiveness (cf. Methods and the SI).
The K-Modes core contains a rich set of short, highly frequent and polysemous words compared to the conformity core. The conformity core contains a more homogeneous set of words, which is crucial for syntactic sentences mixing specific and more general concepts \cite{cancho2001small,pepper2004takes,casas2018polysemy}. The structural k-core has high transitivity, but the conformity core has a more \textit{cliquish} configuration due to higher hub dominance score \cite{yamaguchi2020controlling}. Cliquishness was recently shown to correlate with better recall from memory \cite{valba2021k} due to the concentration of spreading activation in the clique  \cite{kumar2021critical}. These recent studies suggest that the higher cliquishness found here for the conformity core might be beneficial for language processing in toddlers.
The conformity core also displays high values of conductance and cut ratio: this language kernel possess a dense internal structure but it is also strongly connected to the rest of the graph as well, considerably more than the other competitors. In other words, the conformity core is strongly internally connected (more than k-core) and homogeneous with respect to the features (more than k-mode). This higher connectivity might reflect an advantage in accessing and producing items from the language kernel in view of activation spreading models of the mental lexicon \cite{kumar2021critical,kenett2019can,castro2020contributions,levy2021unveiling}.

\noindent {\bf Normative word learning as random walks on \mname}. To investigate how the conformity core and the whole \mname\ structure emerge over time, we adopt a random walk framework. Random walks on cognitive network structures have successfully modelled phenomena like Zipf's law \cite{ferrer2018origins} or semantic priming \cite{griffiths2007google}. Here, we use structure-feature biased random walks to explore normative language learning, as reported in Fig. \ref{fig:fig_2}.

The simplest idea is to limit the walk to network structure only (\textit{Graph Walk 1}). To explore the interplay between topology and features of words, we can weigh network links with the similarity between vectors representing adjacent words (\textit{Graph Walk 2}). Let us consider an example. In Figure \ref{fig:fig_2}, at $t_2$, \textit{Graph Walk 1} should choose to learn either \textit{cat} or \textit{daddy} after the current word \textit{mommy}. Because of network/vectorial similarities, \textit{Graph Walk 2} will select \textit{daddy} as the \textit{next-to-be-learned} word. We can expand the set of next-to-be-learned candidate words: \textit{Graph Walk 3} encodes this parallel word learning process by considering as potential candidates all neighbors of already learned words. With reference to Fig. \ref{fig:fig_2} A, at $t_3$, \textit{Graph Walk 2} can only move to and learn \textit{friend}, while \textit{Graph Walk 3} can also activate and learn \textit{cat} after \textit{mommy}. Focus is given to  considering how these models can predict the assembly over time of: (i) the conformity core, and of (ii) Communicative Development Inventories \cite{fenson2007macarthur} (CDIs), which are commonly used by psycholinguists to measure a child's communicative, receptive and expressive abilities. CDIs are clusters of words from the same semantic category - e.g. a list of words all relative to \textit{time} - and thus represent a portion of the whole vocabulary available to children \cite{macwhinney2014childes}.

\noindent {\bf CDIs are not filled uniformly under normative learning}. In the CHILDES data \cite{macwhinney2014childes}, toddlers are found not to learn CDIs uniformly over time (cf. Fig. \ref{fig:fig_2}, B). This means that some semantic domains of toddlers' lexicon are filled earlier during normative learning. However, the above random walkers do not include information about the semantic category a word belongs to. \textit{Graph Walk 4} proposes a CDI-based similarity integrating information about CDIs' availability and attractiveness.
In the figure, at $t_2$, \textit{Graph Walk 4} moves from \textit{mommy} to \textit{cat}, because \textit{Animal}-CDI is relatively emptier than \textit{People}-CDI, i.e. \textit{People} already contains \textit{mommy}. However, at $t_4$, the model learns \textit{friend} from \textit{daddy}, because the \textit{feature similarity equation} term is stronger than the CDI-based ones (see Methods and SI).

\noindent {\bf Toddler's language kernel rises from CDI density and network/vector dualities.} Figure (\ref{fig:fig_2}, C - left) reports precision and recall in reconstructing the conformity core early on during cognitive development. Performance metrics statistically higher than random learning (significance of 0.05, see SI B) are highlighted with full bars. Non-significant results are visualised as dashed bars.
The normative growth of children's language kernel was captured with a precision higher than random learning only by our most advanced model, combining CDI density, multiplex network structure and feature similarities. This provides strong evidence that semantic spheres and their filling over time provide insights additional to network/vector duality for capturing how early production of syntactically coherent sentences is achieved \cite{hadley2018sentence}. Compared to other CDIs (see next section), our walkers achieved a relatively lower precision in predicting the assembly of the conformity core. This indicates that the language kernel \textit{does not} emerge all at once during early cognitive development, unlike other kernels highlighted in older children \cite{stella2018multiplex}. The emergence of the conformity core is thus a gradual phenomenon, that is not strongly biased by similarities and cannot thus be captured by biased random walks only.

\noindent {\bf Random walks highlight different strategies at work in different CDIs}. Random walks produce word ordering lists that we evaluate with respect to CHILDES normative ordering, i.e. the order in which most children produced words over time (Fig. \ref{fig:fig_2}, B - above). Random learning is used as a baseline to test whether walks considering word topology and feature predict more words as correctly learned over time. See Methods/SI for details of our statistical approach.

Table \ref{table:rw_res} presents a coarse-grained evaluation of the walkers (cf. Discussion). Fig. \ref{fig:fig_2}, B sums up results with respect to CDIs focusing on the very early stage of acquisition, which corresponds to $N=100$ words learned before 21 months \cite{stella2017multiplex}. The selected CDIs are captured differently by the models. CDIs like \textit{People} and \textit{Prepositions} are predicted with higher-than-random precision and recall for all Graph Walk models. \rwD\ precision is slightly better than in the other models. Interestingly, the two most filled CDIs in this stage of acquisition, i.e. \textit{Body} and \textit{Vehicle}, are predicted with high precision but low recall (cf. Methods/SI). This means that the few words predicted are the expected ones, but the models cannot fill the CDIs. \rwC\ precision is higher. Not all CDIs can be predicted in this way, e.g. \textit{Action} and \textit{Household}. Furthermore, model performances for \textit{Household} are not distinguishable from a random learning, i.e. all bars are dotted. The high recall but the low precision of \textit{Action} is poorly relevant: less of 0.1\% of the CDI is covered in this stage of acquisition (however, cf. the SI, where \textit{Action} category is well captured in other stages).

\begin{table}
\begin{adjustbox}{width=1\textwidth}
\begin{tabular}{r|c|c|c} 
     & Accuracy $\vert$ Relevant CDIs & Precision $\vert$ Relevant CDIs & Recall $\vert$ Relevant CDIs \\ 
    
    \parbox[]{4mm}{\multirow{5}{*}{\rotatebox[origin=c]{90}{AoA $<$ 21}}}
    
    Random Learning  & 0.67 $\vert$ || & 0.17 $\vert$ || & 0.19 $\vert$ || \\
    Struct  & 0.70 $\vert$ 0.26 & 0.40 $\vert$ 0.64 & 0.30 $\vert$ 0.58 \\
    MaxSim  & 0.76 $\vert$ 0.26 & 0.37 $\vert$ 0.70 & 0.34 $\vert$ 0.52 \\
    ExtCand  & 0.65 $\vert$ 0.21 & 0.55 $\vert$ 0.76 & 0.30 $\vert$ 0.58 \\ 
    CDI-MaxSim  & 0.75 $\vert$ 0.42 & 0.25 $\vert$ 0.58 &  0.34 $\vert$ 0.47 \\ 
    
    \parbox[]{4mm}{\multirow{5}{*}{\rotatebox[origin=c]{90}{\footnotesize21 $<$ AoA $<$ 23}}}
    
    Random Learning  & 0.71 $\vert$ || & 0.17 $\vert$ || & 0.19 $\vert$ ||  \\
    Struct  & || $\vert$ 0.00 & 0.24 $\vert$ 0.64  & 0.24 $\vert$ 0.71 \\
    MaxSim  & 0.82 $\vert$ 0.36 & 0.28  $\vert$ 0.57 & 0.25 $\vert$ 0.64 \\
    ExtCand  & 0.83 $\vert$ 0.26 & 0.24 $\vert$ 0.42 & 0.24 $\vert$ 0.50 \\ 
    CDI-MaxSim  & 0.66 $\vert$ 0.10 & 0.26 $\vert$ 0.71 & 0.26 $\vert$ 0.71 \\ 
    
    \parbox[]{4mm}{\multirow{5}{*}{\rotatebox[origin=c]{90}{\footnotesize23 $<$ AoA $<$ 24}}}
    
    Random Learning & 0.69 $\vert$ || & 0.17 $\vert$ || & 0.19 $\vert$ || \\
    Struct  & 0.73 $\vert$ 0.21 & 0.19 $\vert$ 0.42 & 0.21 $\vert$ 0.52\\
    MaxSim  & 0.75 $\vert$ 0.36 & 0.17 $\vert$ 0.42 & 0.23 $\vert$ 0.42 \\
    ExtCand  & 0.73 $\vert$ 0.31 & 0.20 $\vert$ 0.21 & 0.21 $\vert$ 0.52\\
    CDI-MaxSim  & 0.69 $\vert$ 0.21 &  0.19 $\vert$ 0.52 & 0.23 $\vert$ 0.52\\ 
    
    \parbox[]{4mm}{\multirow{5}{*}{\rotatebox[origin=c]{90}{\footnotesize 24 $<$ AoA $<$ 26}}}
    
    Random Learning & 0.70 $\vert$ || & 0.17 $\vert$ || & 0.19 $\vert$ ||  \\
    Struct & 0.73 $\vert$ 0.31 & 0.20 $\vert$ 0.44 & 0.22 $\vert$ 0.61 \\
    MaxSim  & 0.72 $\vert$ 0.31 & 0.21 $\vert$ 0.38 & 0.26 $\vert$ 0.44 \\
    ExtCand & 0.71 $\vert$ 0.42 & 0.18 $\vert$ 0.33 & 0.22 $\vert$ 0.50 \\
    CDI-MaxSim & 0.72 $\vert$ 0.31 & 0.23 $\vert$ 0.38 & 0.22 $\vert$ 0.44\\ 
    
    \parbox[]{4mm}{\multirow{5}{*}{\rotatebox[origin=c]{90}{AoA $>$ 26}}}
    
    Random Learning & 0.61 $\vert$ || & 0.24 $\vert$ || & 0.24 $\vert$ || \\
    Struct  & 0.68 $\vert$ 0.78 & 0.32 $\vert$ 0.72 & 0.36 $\vert$ 0.61 \\
    MaxSim  & 0.70 $\vert$ 0.84 & 0.33 $\vert$ 0.77 & 0.35 $\vert$ 0.66\\
    ExtCand  & 0.64 $\vert$ 0.52 & 0.28 $\vert$ 0.44 & 0.29 $\vert$ 0.66\\
    CDI-MaxSim  & 0.79 $\vert$ 0.31 & 0.33 $\vert$ 0.77 & 0.41 $\vert$ 0.66\\ 
\end{tabular}
\end{adjustbox}
\caption{Model performances over each bin of acquisition. \textit{Relevant CDI fraction} is the ratio of statistically significant precision/recall values against a random learning model.}
\label{table:rw_res}
\end{table}

\section{Discussion}

This work introduces a cutting-edge combination of network \cite{collins1975spreading,steyvers2005large,vitevitch2019can} and vector \cite{gunther2019vector,landauer1997solution} aspects of knowledge in the human mind, which historically run in parallel when modelling language and its cognitive processes \cite{castro2020contributions}.

Using data from 1000 toddlers between 18 and 30 months from the CHILDES project \cite{macwhinney2014childes}, our \mname\ network revealed a core of words facilitating word production \cite{pepper2004takes} and invisible to methods based on network structure \cite{stella2017multiplex,stella2018multiplex,holme2005core} or vector similarities only. This core was detected via conformity \cite{citraro2020identifying}, a metric extending assortative mixing estimation in a multi-scale, node-centric fashion. Our numerical experiments identified this core as a set of highly frequent, short, polysemous and well-connected nouns and verbs, i.e. a language kernel containing concepts versatile enough to communicate via basic sentences (cf. \cite{cancho2001small}) and whose access via spreading activation is facilitated by network connectivity \cite{kumar2021critical,castro2020contributions,valba2021k}. Revealing the presence of such a core through our analyses provides for the first time quantitative support of recent empirical findings showing that typical learners can start combining words in basic sentences after 30 months of age \cite{pepper2004takes}. The kernel persisted even when co-occurrences from child-directed speech were ignored (see SI): the conformity core emerged from an interplay between semantic/phonological associations and psycholinguistic norms in the mental lexicon of linguistic knowledge.

To investigate the assembly over time of such a crucial core of linguistic knowledge, we implemented artificial models of word learning as biased random walks over \mname, inspired by past approaches using walkers to investigate the mental lexicon \cite{griffiths2007google,ferrer2018origins}. We found that the conformity core does not emerge suddenly over time, differently from other language kernels modelled as viable component in other studies \cite{stella2018multiplex}. Instead, the conformity core is progressively built in ways that are captured only by combining the network and vector aspects of words together with CDI filling rates. This finding quantitatively stresses that the conformity core - containing building blocks for producing syntactically coherent words - emerges from strategies dependent on semantic categories, which are partly captured by CDIs \cite{macwhinney2014childes}. 

We also used the same random walkers for capturing how different CDIs filled over time through normative learning, giving unprecedented focus \cite{stella2017multiplex} to learning strategies for individual aspects of children's knowledge. In our analyses, different CDIs are found to fill at different times over developmental stages, further emphasizing that language learning is not a uniformly random process. Inventories relative to food and action themes are found to be predicted well by our model, confirming recent independent studies \cite{chang2020adjacent,pomper2019familiar} that these salient familiar themes are crucial for predicting early language acquisition.

Notice also that words in some CDIs might be learned according to context-specific strategies \cite{siow2021exploring,clerkin2017real}, so that a single, general word-learning strategy might not fit all cases. For instance, according to the \textit{Pervasiveness Hypothesis} by Clerkin and colleagues \cite{clerkin2017real}, toddlers would tend to learn earlier words more frequently occurring across several daily contexts. This visual prevalence/occurrence would be crucially missing from CDIs like \textit{Household} or \textit{Action}, which were in fact poorly reproduced by our model. These negative findings indicate the presence of local strategies for learning words in physical settings that are at work in toddlers but missing from the current instance of \mname. 

For inventories like \textit{Body} or \textit{Vehicle}, a combination of network structure and feature similarities corresponded to a significant boost in precision over predictions from random learning. This is quantitative evidence for combining network and vector aspects of the mental lexicon. A further boost in precision was found when the random walker was allowed to backtrack. This indicates that some components of the mental lexicon are not built sequentially, without appending words to the most recent lexical item, as assumed in attachment kernel models \cite{stella2015patterns}, but rather filling gaps in the whole vocabulary available to children, as shown also by other approaches with persistent homology and gap filling \cite{sizemore2018knowledge}.

Interestingly, recency in word acquisition is found to be more a powerful strategy for reconstructing the filling of CDIs like \textit{People} or \textit{Prepositions}, where our most elaborate random walker based on recency beats the back-tracking one. Our quantitative results open the way for further discussion and interpretation in light of psychological studies behind early language learning.

This first conception of \mname\ has some key limitations, which can be addressed in future research. For example, our approach considers only normative learning, i.e. how most children learn words over time \cite{stella2017multiplex}. This learning dynamic might be different from how individual children with different language learning skills might learn words over time \cite{beckage2019network}. Future research should thus test the presence of the language kernel and its time-evolution dynamics in a longitudinal cohort of children. Since the occurrence of the language kernel characterises normative learning in a large population of 1000 and more toddlers \cite{macwhinney2014childes} and it supports the production of early sentences observed in normative talkers \cite{hadley2018sentence}, we expect for the kernel to be present in normative learners but also to be disrupted or incomplete in late talkers \cite{beckage2011small}. If supported by data, then the language kernel revealed here could become a crucial early predictor of delayed language development in young children. Another limitation is that our predictions do not treat learning as the outcome of a statistical process, where words are learned with certain probabilities. Rather we model word learning as a binary learned/not learned process. We chose to follow this approach for model parsimony and indicate the addition of statistical learning \cite{romberg2010statistical} within the \mname\ framework as an exciting future research direction. Future enhancements of random-walk models should account also for distinctiveness in addition to similarity. The recent work by Siew \cite{siew2021global} indicates that global feature distinctiveness, i.e. how many different semantic features are possessed by a word, correlates with earlier acquisition. Hence, random walkers accounting for switches between distinctiveness and similarity might enhance prediction results and represent an exciting future research direction. Another important approach for future research might be casting language acquisition as a percolation problem, which has been explored in feature-rich networks only recently \cite{artime2021percolation}. An important limitation of our study is that it adopts CDIs for modelling language learning, however these inventories are not grounded in theories from cognitive psychology \cite{fenson2007macarthur} but were rather created \textit{ad-hoc} by psycholinguists. Future instances of \mname\ networks should rely on word learning data that is more representative across semantic and syntactic categories.

\section{Methods}

\noindent {\bf Multiplex Layers}.
We modelled word learning as a cognitive process acting on a mental representation of linguistic knowledge. Structure in this representation is given by a multiplex lexical network, where nodes represent words that are replicated and connected across different semantic and phonological levels of the network  \cite{stella2017multiplex}.

Only layers of relevance for word learning acquisition were considered \cite{stella2017multiplex}, namely: (i) free associations, indicating memory recall patterns between words from semantic memory \cite{de2019small}, (ii) co-occurrences in child-directed speech \cite{stella2017multiplex,macwhinney2014childes}, (iii) feature-sharing norms, indicating which concepts shared at least one semantic feature from the McRae dataset \cite{mcrae2005semantic} and (iv) phonological similarities \cite{vitevitch2008can}, representing which words differed by the addition/substitution/deletion of one phoneme only. Hills and colleagues showed that the words with larger degrees in free association networks were also more likely to be acquired at earlier ages, a phenomenon known also as \textit{lure of the associates} (cf. also \cite{hills2018filling}). A subsequent study by Carlson and colleagues \cite{carlson2014children} found a similar effect also in phonological networks built from child-directed speech \cite{vitevitch2008can}. Investigations of co-occurrence and feature sharing networks by Beckage and Colunga reported that highly connected words were distinct trademarks of early word production in typical talkers \cite{beckage2016language}. Importantly, these four aspects of knowledge in the human mind produced network representations that were irreducible \cite{stella2017multiplex}. Layers represented different connectivity patterns among words and could thus not be aggregated or erased without decreasing structural information about the system in terms of Von Neumann graph entropy.

\noindent {\bf Normative age of acquisition}.
Network models of language acquisition often use normative datasets that follow the development of language production in toddlers \cite{hills2018filling}. The most prominent data source is CHILDES (Child Language Data Exchange System), a multi-language corpus of the TalkBank system established by MacWhinney and Snow, storing data about language acquisition in toddlers between age 16 and 36 months \cite{macwhinney2014childes}. 

We used CHILDES data to rank words in the order they are learned by most English toddlers. By considering the fraction of children producing a certain word in a given month, within each month, words were assigned a production probability. Month after month, a rank in descending order of production probability was constructed as a proxy for normative learning of most toddlers, as done in previous studies \cite{hills2009longitudinal,beckage2016language,stella2017multiplex}.

\noindent {\bf Features}.
This study selected word features shown in previous research to influence early language acquisition, namely frequency in child-directed speech \cite{macwhinney2014childes,stella2017multiplex}, word length \cite{hills2009longitudinal,beckage2019network} and polysemy \cite{casas2018polysemy}. Polysemy scores indicated the numbers of meanings relative to a given word in WordNet \cite{miller1998wordnet}, a proxy to word polysemy successfully used in quantitative studies of early word learning \cite{stella2017multiplex}. Due to the highly-skewed distribution of variables (e.g., Zipf's law for word frequency \cite{zipf2016human}), we regularised data by recasting it from numerical to categorical, as to avoid biases in computing conformity \cite{rossetti2021conformity}. We grouped each variable into discrete bins, fine tuning bin boundaries so as to obtain non-empty bins featuring the same order of magnitude of entries. This fine-tuning led to splitting words in quintiles for both word frequency and polysemy and in tertiles for length.

\noindent {\bf Conformity}. We characterise the interplay between structure and features through conformity \cite{rossetti2021conformity}, which estimates the mixing patterns of nodes in a feature-rich network, i.e. a categorical node-attributed network. This measure can find heterogeneous behaviour among all nodes of a network. Conformity enables a multi-scale strategy by leveraging node distances for computing label-similarities between a target node and other nodes. A distance damping parameter $\alpha$ is needed for decreasing the impact of label-similarities over longer network distances between the target node and its connected neighbors. Based on previous investigations \cite{rossetti2021conformity}, we adopt a value of $\alpha=2$ giving more emphasis to closer neighbours in a given network topology. See the SI or \cite{rossetti2021conformity} for a formal description of the measure and the motivation behind its choice in this work.

When analysing conformity, we need to test whether the measured values are a trivial consequence of structural (or attributive) patterns or rather come from a non-trivial interplay between the two. To characterise this, we resort to two null models: (i) random re-shuffling the node attribute labels while maintaining network topology (\textit{Null Model 1}, Fig. \ref{fig:fig_1}, D, \cite{stella2018multiplex}), and (ii) randomly rewiring of links while preserving the node degree and attribute labels (\textit{Null Model 2}, Fig. 1, \ref{fig:fig_1} D). In other words, let us consider this question: Are two labels at the endpoint of an edge significant for the distribution of conformity or can we observe similar patterns by randomly rewiring the attributive or structural model components? While rewiring labels or connectivity patterns, respectively, we keep the other component fixed. For building \textit{Null Model 1}, a random label permutation is enough to disrupt correlations between structure and features.
For building \textit{Null Model 2}, we used a configuration model \cite{molloy2011critical} to obtain a degree preserving graph randomisation, that is, given $N$ nodes and any arbitrary degree sequence $\{k_i\} = (k_1, k_2, k_N)$, we place $k_i$ stubs on each node $i$ in the graph; then we match each stub with another one until all stubs are matched.
The conformity distributions of the null models in Fig. \ref{fig:fig_1}, D refer to the average node scores from 100 randomised instances of \mname\ network.

All conformity distributions are analysed through kernel density estimates (KDEs) and ridgelines (\ref{fig:fig_1}, D); in particular, these last ones get a better picture of mixing heterogeneity between the class labels on the original toddlers' lexicon.

\noindent {\bf Core: Definition and Evaluation}.
For finding a potential language core, we model each word as a vector of conformity scores. 
This results in a vector space where classic clustering algorithms as K-Means \cite{macqueen1967some} can be run.
We reveal a relevant set of words among the six optimal clusters identified by K-Means through the elbow method.
The SI provides methodological details about this configuration.

A set of several quality functions are proposed to characterise the language core.
We focus on modularity, conductance, cut ratio, internal edge density, hub dominance and transitivity \cite{newman2018networks}. Modularity, conductance and cut ratio focus on the links within and outside a community: They measure how well-separated a cluster is from the rest of the network. Edge density, transitivity and hub dominance characterise the internal structure of the core. In particular, transitivity and hub dominance characterise it in terms of triadic closure and \textit{cliquishness} level, i.e. the creation of subgraphs where each node is fully connected to others. See the SI for their formal description. All in all, these network metrics are used to characterise the structure of the different cores found via conformity (in \mname), via core-detection on the network structure only \cite{holme2005core} and via K-Modes on feature embeddings only \cite{huang1997clustering}. Notice that these measures, combined, provide info about the distinctiveness and connectedness of a given component/cluster in a network. 

\noindent {\bf Graph Walks}.
We aim to model early word acquisition by progressively combining the network and vector components of \mname\. To achieve this goal, the core idea is to generate a word rank that is progressively filled according to the different graph walk strategies, each one incorporating specific assumptions. 
In this work we compare four alternative random walk models each one having a unique rationale on how to weigh links and/or to determine the set of candidates for the next to-be-learned word. In particular:

\begin{itemize}
    \item \rwA\ (Graph Walk 1): Words are connected by unweighted links, hence the next word is chosen according to the underlying structure only. Similarly, the set of candidates is chosen from the adjacent neighborhood of the current word;
    \item \rwB\ (Graph Walk 2): Edges are weighted according to the pairwise similarity between nodes' features. Jaccard similarity is used (cf. SI), and  frequency, length and polysemy are all considered. The same strategy of \rwA\ is used for the set of candidates;
    \item \rwC\ (Graph Walk 3): The same strategy of \rwB\ is used for weighing links; the set of candidates is chosen from the adjacent neighborhood of all the words already learned;
    \item \rwD\ (Graph Walk 4): Links are weighted according to a CDI-based pairwise similarity between the attributes of nodes as well as the availability and attractiveness (cf. SI), and it needs to be updated at each iteration. The same strategy of \rwA\ and \rwB\ is used for the set of candidates.
\end{itemize}

\rwA\ and \rwB\ are biased random walks considering, respectively, topology or similarity between words (i.e., the network structure or the vector space) while \rwC\ and \rwD\ aim for a more holistic approach.

\rwC\ visit strategy is designed to mime non-sequential word learning in children (cf. \cite{beckage2019network}), where the word acquired at step $t+1$ could be similar to any word already learned before, thus enabling an interplay between exploration and exploitation of CDIs. 
When the last word determines the topology of similar candidates for the next acquisition step, resembling a Markovian process \cite{hills2009longitudinal}, the walker possesses a bias to remain within the same CDI. By considering as to-be-learned candidates all previously learned words, the walker has a chance of backtracking and acquiring more words within the CDI sharing tightly similar concepts.

\rwD, the CDI-based model relies on pairwise similarity between two words modulated by additional information on the filling of CDIs they belong to. For additional details and a formal description of the pairwise similarity function adopted refer to the SI.

\noindent {\bf Graph Walk Evaluation}. Accuracy, precision and recall are used to evaluate the goodness of ranks' prediction, as commonly done in statistics and machine learning. Accuracy is defined as the number of correct predictions, i.e. true positives or TP, divided by the total number of predictions. In this domain, TPs are words belonging to a CDI that are learned by a random walker in a specific bin of age of acquisition. Precision is the fraction of relevant elements among all the retrieved ones including non-relevant elements, i.e. false positives or FP. In this domain, FPs are words that fill a CDI as expected in a particular age of acquisition bin, but they are not the exact same words considered in normative learning. For instance, \textit{dog} might contribute to increase FPs because it belongs to the \textit{Animals} CDI but the normative learning contemplated \textit{cat} instead of \textit{dog}. Finally, recall is the fraction of relevant elements that are retrieved. Missing relevant elements (false negatives or FNs) are CDI's words that are not retrieved by a random walker in a particular bin of age of acquisition. 
The above definitions imply that there can be predictions with high recall and low precision, because there are many words that satisfy the semantic category roughly represented by the CDI (e.g. guessing as learned names of animals) but different from the specific words learned during normative acquisition (e.g. other names of animals). This interplay spans from the specific characterisation of random-walk predictions and it is accounted for in the Results and Discussion sections. 
See the SI for a complete formalization of the measures, and toy examples.


\clearpage


{\Huge
\textbf\newline{Appendix: Supporting Information}
}

\appendix

\section{Conformity}

\subsection{Definition}

Conformity provides a multi-scale strategy to estimate local homophily in complex networks, overcoming classic measures as Newman's assortativity \cite{newman2003mixing} that only produce a global, averaged score.
Conformity is not the only multi-scale strategy in the literature. Some valuable variants of local Newman's assortativity also exist \cite{peel2018multiscale}.
The reason behind the choice of conformity rather than other measures is because conformity provides node similarities grounded in the real distances between nodes. Other approaches, for instance, only leverage random walks as a proxy of information about paths of all possible lengths \cite{peel2018multiscale, bassolas2021first}.

To the best of our knowledge, no other works have used node-centric homophily estimation in a mental lexicon. These measurements were typically applied in social network analysis \cite{mcpherson2001birds} or mobility data \cite{bokanyi2021universal}.

In the following, we report a concise description of conformity \cite{rossetti2021conformity}. 
Given a node-attributed graph $G=(V,E,A)$, where $V$ is the set of nodes, $E$ the set of edges, and $A$ the set of categorical node attributes, we calculate for a node $u \in V$ the conformity score $\psi(u, \alpha)$ with respect to an attribute $l \in A$.
The damping parameter $\alpha$ allows to decrease the impact of label-similarities over longer network distances between the target node and all the reachable neighbors.
To define conformity formally, we need a couple of support functions, namely the indicator $I_{u,v}$ 

\begin{equation}
    I_{u,v} = 
\left\{
    \begin{array}{ll}
        1  & \mbox{if } l_u=l_v \\
        -1 & \mbox{otherwise},
    \end{array}
\right.
\label{eq:Iuv}
\end{equation}

that compares the attribute values of two nodes, and the similarity function $f_{u,l_u}$

\begin{equation}
    f_{u,l_u} = \frac{|\{v|v\in \Gamma_u \land l_u=l_v\}|}{|\Gamma_u|},
\end{equation}

that computes the ratio of $u$'s first-order neighbors that share the same attribute value $l_u$.
Thus, given a real number $\alpha$ in $[0, +\infty)$, conformity of node $u\in V$ is defined as in the following:

\begin{equation}
    \psi(u,\alpha) = \frac{\sum_{d\in D} \frac{\sum_{v \in N_{u,d}} I_{u,v} f_{v,l_v}}{|N_{u,d}| d^\alpha}}{\sum_{d\in D} d^{-\alpha}},
    \label{eq:conformity}
\end{equation}

where $D=max(\{dist(i,j)|i,j \in V\})$, i.e. the maximum distance among all node pairs.
The computed score is normalized to ensure that conformity lies in the range $[-1, 1]$.

\subsection{Comments on null models}

Some class labels exhibit more assortative mixing than others, e.g. shortest words in conformity by length or most frequent words in conformity by frequency (cf. Results). 
While reshuffling node labels or rewiring links, we intend to observe whether similar distributions can emerge trivially from random label permutations or random link configurations.
The heterogeneous distributions do not emerge while measuring conformity on the ensemble of networks obtained from the two randomisation processes (cf. Results and Methods).
In conformity by frequency and by length, null models distributions are mainly disassortative among all classes; hence, the \textit{anomalous} behaviour of most frequent and shortest words is flattened by both randomisation processes.
Conformity distribution with respect to polysemy slightly differs from the other two attributes. In particular, the null model that rewires links shows a bi-modal distribution.
The explanation behind this behaviour could be similar to the explanation used to describe the quasi uniformly mixed pattern of polysemous words (cf. Results): ambiguous and unambiguous words must link in non-trivial patterns that are harder to break while randomising node connectivity.

\subsection{Conformity vector space}

Multi-dimensional conformity information is used in the language core analysis for finding a relevant set of words in language acquisition.
We model each node as its vector of conformity scores, where conformity by frequency, by length and by polysemy are the vector components.
This allows to build a vector space where classic clustering analysis can be performed.
The difference between a clustering method on the features only (cf. K-Modes \cite{huang1997clustering}, Results) is that using vector of conformity scores we integrate structure-feature relationships, thus we aim to group words having similar mixing patterns across the features.

An optimal K-Means \cite{macqueen1967some} instance is used to cluster words. For selecting the optimal number of clusters $k$, we leverage the elbow method, namely plotting the sum of squared errors in function of $k$ and determining the point of inflection of the curve.
$k=6$ is identified as the optimal point and chosen as the number of centroids to initialize the algorithm.

\section{Core Evaluation}

We report here a description of the quality functions used for characterising the language core(s).
All the measures are implemented in the CDLib library \cite{rossetti2019cdlib}, and other detailed information can be found in the library documentation.
Let $G=(V,E)$ be a graph with $v \in V$ and $e \in E$, and $C$ a partition of $G$ with $c \in C$, with $c$ composed by a subset of $V$ and a subset of $E$.
We aim to characterise a cluster/community/core \textit{c} with the following measures:

\begin{itemize}
    \item \textit{Conductance}: the fraction of total edge volume that points outside the community:
    
    \begin{center}
        $conductance(c) = \frac{|c|}{2|e_c| + |c|}$,
    \end{center}
    
    where $|c|$ is the cardinality of the community, namely the number of community nodes, and $e_c$ the number of community edges;
    
    \item \textit{Edge density}: the internal density of the community set:
    
    \begin{center}
        $density(c) = \frac{|e_c|}{\frac{|c|(|c|-1)}{2}}$;
    \end{center}
    
    \item \textit{Hub dominance}: indicates the ratio of the degree of the most connected node in a community with respect to the theoretically maximal degree within the community, namely
    
    \begin{center}
        $Hub\_dom(c) = \left\{
        \begin{array}{ll}
                1  & \mbox{iff} |c|=1 \\
                    \frac{max_{v \in c} k_v}{|c|-1} & \mbox{otherwise},
                \end{array}
            \right. $,
    \end{center}
    
    where $k_v$ is the degree of node $v$;
    
    \item \textit{Modularity}: measures the strength of the division of a network into sets of well-separated clusters or modules, and it is calculated as the sum of the differences between the fraction of edges that actually fall within a given community and the expected fraction if edges were randomly distributed:
    
    \begin{center} 
    $Q=\frac{1}{(2 e_c)} \sum_{v w}\left[A_{v w}-\frac{k_{v} k_{w}}{(2 e)}\right] \delta\left(c_{v}, c_{w}\right)$,
    \end{center}
    
    where $A_{v,w}$ is the entry of the adjacency matrix for $v,w \in V$, $k_v, k_w$ the degree of $v, w$ and  $\delta\left(c_{v}, c_{w}\right)$ is an indicator function taking value 1 iff $v, w$ belong to the same community, 0 otherwise;
    
    \item \textit{Normalized cut ratio}: is the fraction of existing edges (out of all possible edges) leaving the community:
    
    \begin{center}
        $ cut\_ratio(c) = \frac{|c|}{2|e_c| + |c|} + \frac{|c|}{2(|e| - |e_c|) + |c|} $;
    \end{center}
    
    \item \textit{Transitivity}: is the average clustering coefficient of community nodes with respect to their connection within the community itself:
    
    \begin{center}
        $ CC(c) = \frac{1}{|c|} \sum_{v\in c}\frac{2\Delta}{k_v(k_v - 1)} $
    \end{center}
    where $\Delta$ is the number of triangles including node $v$ in the community $c$.
    
\end{itemize}

\section{Core Analysis}

\subsection{Persistence without layers}

\begin{figure*}[h!]
\centering
  {\includegraphics[scale=0.55]{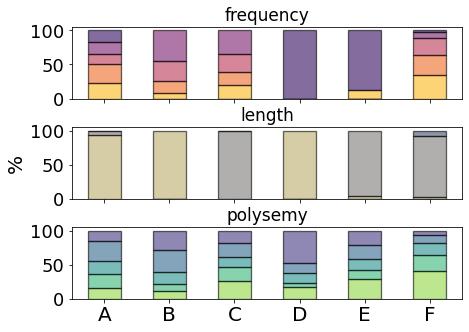}}
  \caption{Conformity vector space characterisation without the co-occurrence layer.}
    \label{fig:fig_nocoocc_}
\end{figure*}

A key result of this work is the identification of a language kernel with interesting structural properties and non-trivial content organisation.
Randomly picked pairs/triads of words from this kernel can build simple, syntactically well-formed sentences (cf. Results).
However, it can be observed that this core can appear just because there is the co-occurrence layer in the multiplex network, i.e. links between concepts co-occurring in child-directed speech.
A strength of the \mname\ model is the possibility to \textit{switch off} a layer from the structural component.
Removing completely a layer also allows us to observe the emergence of an interesting conformity core. 
Fig. \ref{fig:fig_nocoocc_} establishes that a language core emerges even when the 1000 co-occurrences links from the child-directed speech layer are removed.
The language kernel (here, the cluster labeled as D) persists to show homogeneity across all the features.
This gives more strength to the hypothesis that the kernel found with conformity stems from a more broad interplay between semantic and phonological layers.
Finally, for a complete overview of the whole clustering result, the few differences we can notice without the co-occurrence layer is a more homogeneous distribution of word length within the clusters (cf. Fig 1, E), probably due to the removal of long words from the network.

\subsection{Degree assortativity and hierarchical organisation}

\begin{figure*}[h!]
\centering
  \subfloat[]{\includegraphics[scale=0.47]{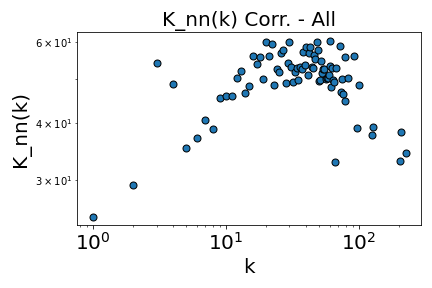}}
  \subfloat[]{\includegraphics[scale=0.47]{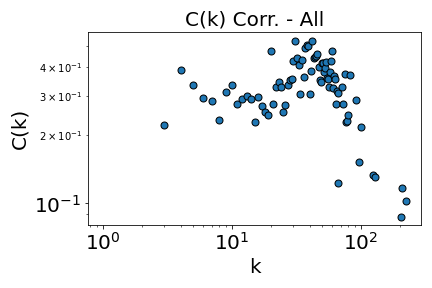}}
  \qquad
  \subfloat[]{\includegraphics[scale=0.47]{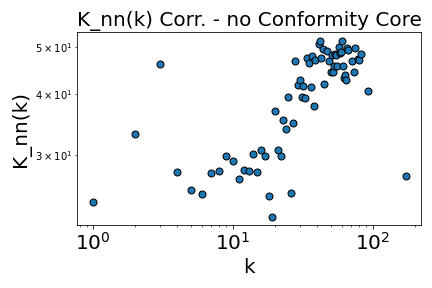}}
  \subfloat[]{\includegraphics[scale=0.47]{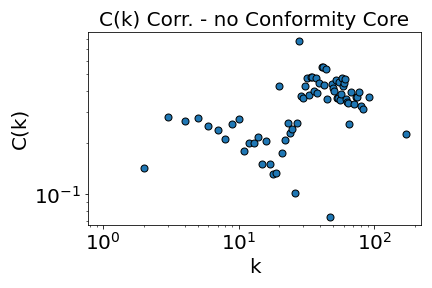}}
  \caption{Degree-degree assortativity and hierarchical organisation in CHILDES mental lexicon studying $Knn(k)$ and $C(k)$ curves, respectively, (a-b) with and (c-d) without the conformity core.}
    \label{fig:fig_curves}
\end{figure*}

We may want to observe whether the removal of the conformity-core can disrupt global characteristics of the network.
Several studies \cite{liu2009statistical, utsumi2015complex, ravasz2003hierarchical} identify complex global properties in the degree-degree assortativity and in the hierarchical organisation of lexical networks, measured through $Knn(k)$ and $C(k)$ curves, respectively.
$Knn(k)$ curves show the average degree of neighbors of nodes with degree $k$ \cite{liu2009statistical}.
Similarly, $C(k)$ curves show the average clustering coefficient of nodes with degree $k$ \cite{ravasz2003hierarchical}.
If $Knn(k)$ increases with $k$, the network behaviour is assortative by degree; if $Knn(k)$ decreases with $k$, the network behaviour is disassortative.
If $C(k)$ decreases with $k$, the network exhibits a hierarchical organisation, otherwise the network does not present this characteristic.
Fig. \ref{fig:fig_curves} shows that $Knn(k)$ and $C(k)$ curves drastically change when removing the language kernel found with conformity.
The  network is highly disassortative and hierarchically organised when all nodes are present, but switches to a highly degree-assortative behaviour and splits into two $C(k)$ branches when the core is removed ($r_{Knn} = -0.27$ with $N=V$, $r_{Knn} = 0.55$ with $N=V-V_{core}$; $r_{C} = -0.33$ with $N=V$, $r_{C} = 0.32$ with $N=V-V_{core}$).
Both disassortativity and hierarchies may spawn from super-general concepts being embedded in the core. These nodes may attract many more links than other nodes, forming a hub-node structure that creates degree disassortativity and implies that the neighbors of hubs are not linked to each other.
We suggest that these findings are coherent with the intra-cluster and inter-clusters analysis to characterise the core (cf. Results). The core is internally structured as a tight \textit{clique}, i.e. it presents high transitivity and hub dominance values. Nevertheless, high conductance/cut-ratio values indicate that the core is highly connected to the rest of the graph.
Thus, removing this set of words disrupts the global disassortative and hierarchical organisation of the system, i.e. the underlying structure that guarantees the system connectivity.
Further analyses are needed if we aim to interpret these results from a cognitive perspective.
Despite some relevant exceptions \cite{van2015examining}, the lack of a compact set of studies on the degree-disassortative behaviour and the hierarchical organisation of lexical networks makes it difficult to shed light on the underlying cognitive phenomena structuring these non-trivial topological patterns.

\subsection{Comparison with other cores}
A language kernel is defined as relatively small set of words enabling the creation of simple yet general and frequently used sentences, thus facilitating early communication \cite{cancho2001small}.
This kernel was identified in the cluster found via conformity. Words in this kernel form simple and syntactically structured sentences, and are heterogeneous in part of speech composition.
Conversely, the kernel found via k-core network decomposition \cite{holme2005core} is unable to form syntactically coherent bigrams/trigrams (cf. Results, Fig. 1, F).
For a more robust investigation, we report additional analysis here, including in the discussion the kernel found via K-Modes clustering as well \cite{huang1997clustering}.

We focus on the whole internal content organisation of the three cores, regardless of their underlying structure, that we characterized up to now. We aim to compute the ratios of the internal syntactic coherence for the cores.
From each core we extract all the possible bigrams.
This approach considers each link from the complete subgraph made of core-words, and allows us to count the frequency of each part of a speech pair.
We find that the most frequent bigram in the structural-based and attribute-based cores is the \textit{noun-noun} pair, 0.42\% and 0.20\%, respectively.
Conversely, the complete subgraph from the conformity-core continues to present a more heterogeneous part of speech composition, where a prominent pair is not observed. In fact, similar frequencies are found for the \textit{verb-noun} (0.1\%), the \textit{noun-adjective} (0.06\%) and the \textit{verb-adjective} (0.05\%) pairs, which are the three most frequent bigrams in the conformity-core.

\section{Graph Walks}

\subsection{Pseudo-code}

Algorithm \ref{alg:rw} introduces a general schema of a graph walk to describe the four proposed variants.
We impose an undirected graph as input, initializing the edge weights as desired (line 1), e.g. all weights are equal to 1 if we want to ignore feature similarity.
Then, the word acquisition ordering is initialized, and a randomly selected node $n$ is added to the rank at $t=1$ (lines 2-3).
The set of word candidates is initialized (line 4) to be filled according to the different strategies of each graph walk.
Once having a starting word and the first set of candidates, the walk starts until the whole dataset is covered, e.g. each node has a position assigned in the rank (line 5).
We iteratively select the new current word by computing the similarity between the current word and the word candidates, choosing the one which maximises similarity (lines 7-9);
we re-compute similarity at each iteration because we might need to update this quantity, e.g. when this last one is based on CDI's availability.
The new current word is added to the rank only if it was not already learned, otherwise it is removed from the set of candidate (lines 10-12).
If the set is empty (line 6), a randomly chosen and still not learned word is added to it for continuing the walk (lines 13-14); this is equivalent to adding an error $\epsilon$, only when it is strictly necessary.

\begin{algorithm}[t!]
\caption{Graph Walk}
\label{alg:rw}
\begin{algorithmic}[1]
\Require Undirected Graph $G=(V,E)$, starting node $n$

\State Initialize weights on $E$

\State Initialize word ordering list $T$

\State Let $n$ be the current visited word and add it to $T$ 

\State Initialize set of word candidates $C$

\While {$len(T) < |V| $}

\If{$C$ is not empty}

\For {each candidate $c$ in $C$}
\State Compute the similarity between $n$ and $c$
\EndFor

\State Let $max(c)$ be the current word

\If{$max(c)$ not in $T$}
\State add $c$ to $T$
\State remove $c$ from $C$
\EndIf

\Else
\State add randomly a not already learned $v \in V$ to $C$
\EndIf

\EndWhile

\State \Return $T$

\end{algorithmic}
\end{algorithm}

\subsubsection{CDI-based similarity}

The CDI-based model relies on pairwise similarity between two words $i$ and $j$ modulated by additional information on the CDIs they belong to, namely $i\in I$ and $j \in J$. Let $A$ be the set of features relative to $i$ and $B$ the features of $j$. The probability for the random walker to acquire $j$ after $i$ is the factorisation of three terms:

\begin{center}
    $ P_{i \rightarrow j} =
            \left\{
                \begin{array}{ll}
                    J(i,j)  & \mbox{iff} g(cdi(j)) \cdot h(i,cdi(j)) = 0  \\
                    J(i,j) \cdot g(j) \cdot h(i,cdi(j)) & \mbox{otherwise},
                \end{array}
            \right.$
\end{center}

The first term in $P_{i \rightarrow j}$ accounts for the similarity between sets $A$ and $B$ quantified via the Jaccard index, namely $sim(A,B)=|A  \cap B|/|A \cup B|$ or the ratio of elements common to both sets of features $A$ and $B$. The second term is the target CDI availability, namely the amount of words still available for acquisition in the CDI containing target word $j$:

\begin{center}
    $g(j) = 1 - \frac{|\{w \in J | is active \}|}{|J|}.$
\end{center}

The more words are available for acquisition in the target CDI $J$, the higher $g(j)$ and thus the probability for the random walker to move to $j\in J$. The third term is the CDI attractiveness:

\begin{center}
    $h(i, J) = \frac{|\{j \in \Gamma_i | j \in J \}|}{|\Gamma_i|}$,
\end{center}

where $\Gamma_i$ is the set of adjacent neighbors of the word available for acquisition $i$. The more $i$'s neighbors are within the target CDI $J$, the higher $h(i,J)$ and thus the probability for the next word to be \textit{attracted} in the CDI where many neighbors are already present.

\subsection{Evaluation}

\begin{figure*}[h!]
\centering
  \subfloat[]{\includegraphics[scale=0.38]{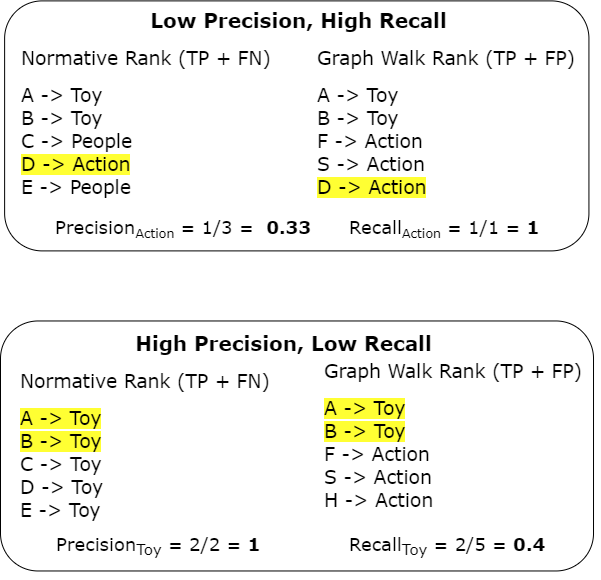}}
  \subfloat[]{\includegraphics[scale=0.35]{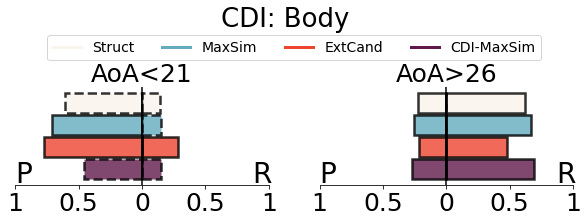}}
  \caption{(a): Toy example focusing on the meaning of precision and recall of a CDI; (b): Precision and recall of \textit{Body} CDI at two different stages of acquisition, i.e., AoA<21 months, namely the first 100 learned words (N=100), and AoA>26, namely the last 129 learned words according to CHILDES dataset.}
    \label{fig:toy_pr_and_body}
\end{figure*}

Accuracy, precision and recall evaluate the performances of the random walks (cf. Results and Methods). 
The measures are built upon the confusion matrix of predictions, i.e. a matrix containing the number of correct predictions, namely true positives (TPs) and true negatives (TNs), and the number of incorrect predictions, namely false positives (FPs) and false negatives (FNs).
Contextualizing these concepts in this domain, TPs are CDI's words correctly learned by a random walker in a selected AoA bin, while TNs are all other words that a graph walk correctly predict as not belonging to a CDI in that AoA bin; FPs are words that fill a CDI as expected, but they are not the exact same words considered in normative learning, while FNs are CDI's words that are not retrieved in that AoA bin.

Formally, accuracy is the number of TPs divided by the total number of predictions:

\begin{center}
    $Accuracy(CDI, AoA) = \frac{TP}{TP+TN+FP+FN}$.
\end{center}

Accuracy can answer poorly to some questions as \textit{how many (expected) CDI's words a graph walk can retrieve in a specific AoA bin?}
Precision and recall address this better.
Hence, precision is the fraction of relevant elements among the retrieved ones,

\begin{center}
    $Precision(CDI, AoA) = \frac{TP}{TP+FP}$,
\end{center}

while recall is the fraction of relevant elements that are retrieved,

\begin{center}
    $Recall(CDI, AoA) = \frac{TP}{TP+FN}$.
\end{center}

\noindent {\bf Example n. 1}. We aim to evaluate the performances of a random walk focusing on how the model is filling the \textit{Animal}-CDI at a very early stage of acquisition, i.e. considering the words learned before 21 months.
$Recall(Animal, <21m)$ increases whatever animal-related word the model retrieves, e.g. \textit{dog} and \textit{frog}, but $Precision(Animal, <21m)$ does not increase if \textit{frog} is not learned before 21 months.
FPs as \textit{frog} are non relevant words;
moreover, the model can miss FNs as \textit{cat}, i.e. relevant words learned before 21 months.

\noindent {\bf Example n. 2}. Fig. \ref{fig:toy_pr_and_body} (a) focuses on two possible extremes, i.e. when precision is low but recall is high (above), or precision is high but recall is low (below).
In the first case, the normative learning contains only one \textit{Action} word in the sliced AoA bin of five words, but the graph walk retrieves three \textit{Action} words.
Recall is maximised, i.e. the expected word D is retrieved; however, FPs as F and S decrease precision.
In the second case, \textit{Toy} words only fill the sliced normative bin, but the graph walk correctly predicts two expected words out of five.
Precision is maximised, i.e. A and B are expected words; however, FNs as C, D and E decrease recall.

\begin{figure*}[h!]
\centering
  {\includegraphics[scale=0.4]{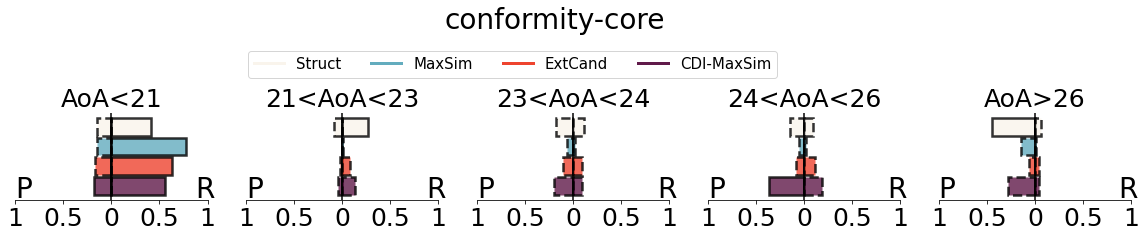}}
  \caption{Precision-recall evaluation of the core over all bins of age of acquisition.}
    \label{fig:pr_core}
\end{figure*}

\begin{figure*}[h!]
\centering
  {\includegraphics[width=\textwidth,height=\textheight, 
  keepaspectratio]{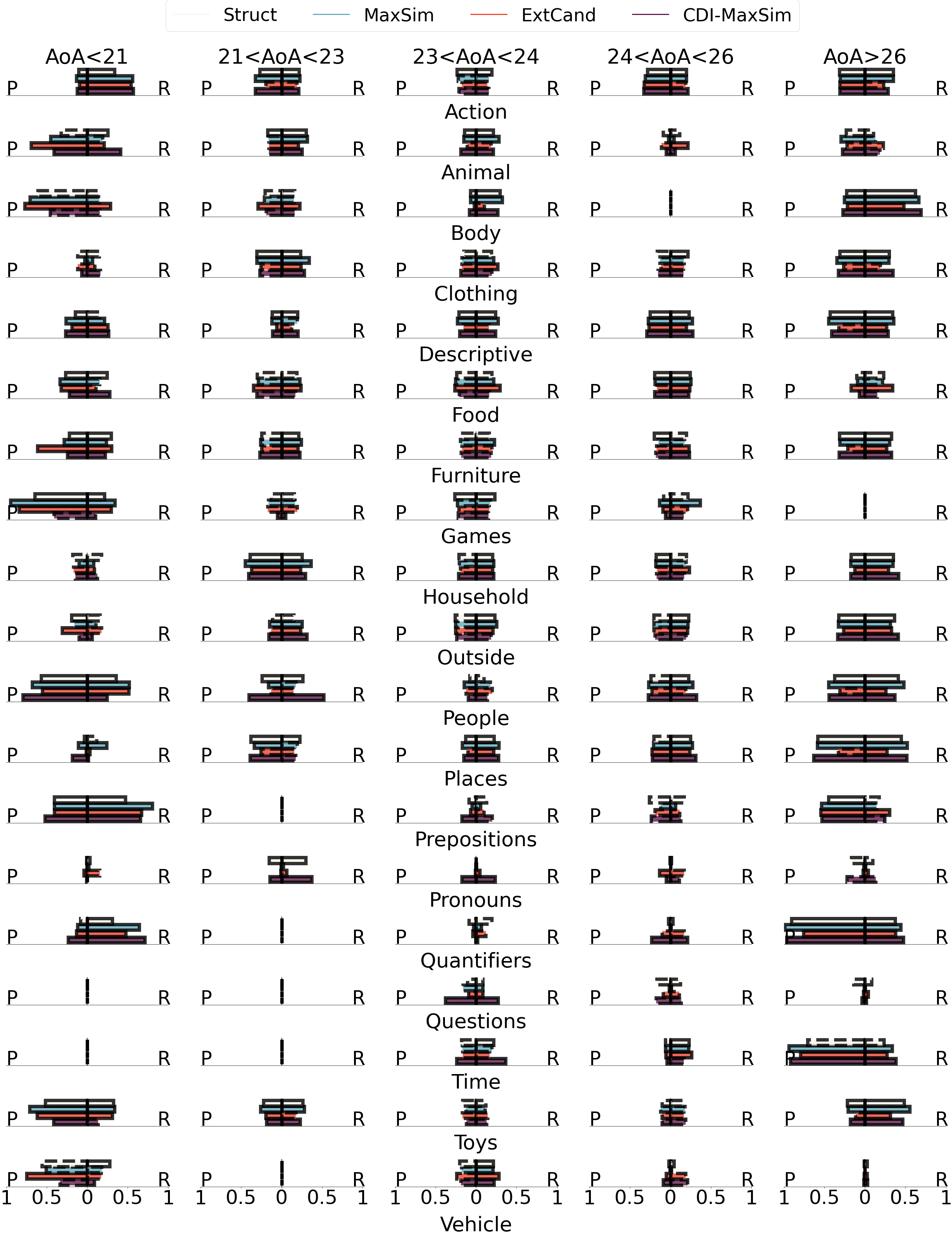}}
  \caption{Precison-recall evaluation of all CDIs over all bins of age of acquisition.}
    \label{fig:pr_all}
\end{figure*}

Fig. \ref{fig:toy_pr_and_body} (b) sums up a real example on the CHILDES dataset.
Fig. \ref{fig:pr_core} reports the precision-recall bars of the conformity-core for each bin of age of acquisition (cf. Results, focus on the core words learned before 21 months only).
Fig. \ref{fig:pr_all} reports the complete precision-recall bars of each CDI and age of acquisition (cf. Results, focus on \textit{Action}, \textit{Body}, \textit{Household}, \textit{People}, \textit{Prepositions} and \textit{Vehicle} CDIs only, acquired before 21 months).

\subsubsection{Random learning}
We compare the graph walk performances against a learning model that assigns to all words a position in the rank randomly, regardless of any type of structural, vectorial or CDI-based information (cf. Results).
In each precision-recall plot, solid bars specify whether the values are statistically significant with respect to this random word assignment.
We use the following z-score for the test: 

\begin{center}
    $z=\frac{M_1-M_2}{\sqrt{\frac{\sigma_1^2}{n}+\frac{\sigma_2^2}{n}}}$
\end{center}

where $M_1$ is the mean precision/recall value from $n$ runs of the walk model, with $\sigma_1$ standard deviations, and $M_2$ is the mean precision/recall value from $n$ runs of the random learning model, with $\sigma_2$ standard deviations.
Thus, dotted bars indicate values that are not statistically different from the random learning distribution ($z>0.05$ or precision/recall higher in the random learning than in the random walk model).

\section*{Acknowledgments}
This work is supported by the European Union – Horizon 2020 Program under the scheme “INFRAIA-01-2018-2019 – Integrating Activities for Advanced Communities”, Grant Agreement n.871042, “SoBigData++: European Integrated Infrastructure for Social Mining and Big Data Analytics” (http://www.sobigdata.eu).

\printbibliography

@article{citraro2020identifying,
  title={Identifying and exploiting homogeneous communities in labeled networks},
  author={Citraro, Salvatore and Rossetti, Giulio},
  journal={Applied Network Science},
  volume={5},
  number={1},
  pages={1--20},
  year={2020},
  publisher={SpringerOpen}
}

@article{rossetti2021conformity,
  title={Conformity: A path-aware homophily measure for node-attributed networks},
  author={Rossetti, Giulio and Citraro, Salvatore and Milli, Letizia},
  journal={IEEE Intelligent Systems},
  volume={36},
  number={1},
  pages={25--34},
  year={2021},
  publisher={IEEE}
}

@inproceedings{huang1997clustering,
  title={Clustering large data sets with mixed numeric and categorical values},
  author={Huang, Zhexue},
  booktitle={Proceedings of the 1st pacific-asia conference on knowledge discovery and data mining,(PAKDD)},
  pages={21--34},
  year={1997},
  organization={Citeseer}
}

@inproceedings{macqueen1967some,
  title={Some methods for classification and analysis of multivariate observations},
  author={MacQueen, James and others},
  booktitle={Proceedings of the fifth Berkeley symposium on mathematical statistics and probability},
  volume={1},
  number={14},
  pages={281--297},
  year={1967},
  organization={Oakland, CA, USA}
}

@incollection{vitevitch2019can,
  title={Can network science connect mind, brain, and behavior?},
  author={Vitevitch, Michael S},
  booktitle={Network science in cognitive psychology},
  pages={184--197},
  year={2019},
  publisher={Routledge}
}

@article{vitevitch2008can,
  title={What can graph theory tell us about word learning and lexical retrieval?},
  author={Vitevitch, Michael S},
  year={2008},
  publisher={ASHA}
}

@inproceedings{vitevitch2014using,
  title={Using complex networks to understand the mental lexicon},
  author={Vitevitch, Michael S and Goldstein, Rutherford and Siew, Cynthia SQ and Castro, Nichol},
  booktitle={Yearbook of the Poznan Linguistic Meeting},
  volume={1},
  number={1},
  pages={119--138},
  year={2014},
  organization={Sciendo}
}

@article{siew2019cognitive,
  title={Cognitive network science: A review of research on cognition through the lens of network representations, processes, and dynamics},
  author={Siew, Cynthia SQ and Wulff, Dirk U and Beckage, Nicole M and Kenett, Yoed N},
  journal={Complexity},
  volume={2019},
  publisher={Hindawi}
}

@article{castro2020contributions,
  title={Contributions of modern network science to the cognitive sciences: revisiting research spirals of representation and process},
  author={Castro, Nichol and Siew, Cynthia SQ},
  journal={Proceedings of the Royal Society A},
  volume={476},
  number={2238},
  pages={20190825},
  year={2020},
  publisher={The Royal Society Publishing}
}

@article{kenett2019can,
  title={What can quantitative measures of semantic distance tell us about creativity?},
  author={Kenett, Yoed N},
  journal={Current Opinion in Behavioral Sciences},
  volume={27},
  pages={11--16},
  year={2019},
  publisher={Elsevier}
}

@article{gunther2019vector,
  title={Vector-space models of semantic representation from a cognitive perspective: A discussion of common misconceptions},
  author={G{\"u}nther, Fritz and Rinaldi, Luca and Marelli, Marco},
  journal={Perspectives on Psychological Science},
  volume={14},
  number={6},
  pages={1006--1033},
  year={2019},
  publisher={Sage Publications Sage CA: Los Angeles, CA}
}

@article{landauer1997solution,
  title={A solution to Plato's problem: The latent semantic analysis theory of acquisition, induction, and representation of knowledge.},
  author={Landauer, Thomas K and Dumais, Susan T},
  journal={Psychological review},
  volume={104},
  number={2},
  pages={211},
  year={1997},
  publisher={American Psychological Association}
}

@article{lund1996producing,
  title={Producing high-dimensional semantic spaces from lexical co-occurrence},
  author={Lund, Kevin and Burgess, Curt},
  journal={Behavior research methods, instruments, \& computers},
  volume={28},
  number={2},
  pages={203--208},
  year={1996},
  publisher={Springer}
}

@article{quillian1967word,
  title={Word concepts: A theory and simulation of some basic semantic capabilities},
  author={Quillian, M Ross},
  journal={Behavioral science},
  volume={12},
  number={5},
  pages={410--430},
  year={1967},
  publisher={Wiley Online Library}
}

@article{collins1975spreading,
  title={A spreading-activation theory of semantic processing.},
  author={Collins, Allan M and Loftus, Elizabeth F},
  journal={Psychological review},
  volume={82},
  number={6},
  pages={407},
  year={1975},
  publisher={American Psychological Association}
}

@incollection{beckage2016language,
  title={Language networks as models of cognition: Understanding cognition through language},
  author={Beckage, Nicole M and Colunga, Eliana},
  booktitle={Towards a Theoretical Framework for Analyzing Complex Linguistic Networks},
  pages={3--28},
  year={2016},
  publisher={Springer}
}

@book{macwhinney2014childes,
  title={The CHILDES project: Tools for analyzing talk, Volume II: The database},
  author={MacWhinney, Brian},
  year={2014},
  publisher={Psychology Press}
}

@article{carlson2014children,
  title={How children explore the phonological network in child-directed speech: A survival analysis of children’s first word productions},
  author={Carlson, Matthew T and Sonderegger, Morgan and Bane, Max},
  journal={Journal of memory and language},
  volume={75},
  pages={159--180},
  year={2014},
  publisher={Elsevier}
}

@article{hills2018filling,
  title={Filling gaps in early word learning},
  author={Hills, Thomas T and Siew, Cynthia SQ},
  journal={Nature Human Behaviour},
  volume={2},
  number={9},
  pages={622},
  year={2018},
  publisher={Nature Publishing Group}
}

@article{stella2017multiplex,
  title={Multiplex lexical networks reveal patterns in early word acquisition in children},
  author={Stella, Massimo and Beckage, Nicole M and Brede, Markus},
  journal={Scientific Reports},
  volume={7},
  pages={46730},
  year={2017},
  publisher={Nature Publishing Group}
}

@book{pepper2004takes,
  title={It takes two to talk: A practical guide for parents of children with language delays},
  author={Pepper, Jan and Weitzman, Elaine},
  year={2004},
  publisher={The Hanen Centre}
}

@article{valba2021k,
  title={K-clique percolation in free association networks. The mechanism behind the $7 \pm 2$ law?},
  author={Valba, Olga and Gorsky, Alexander},
  journal={arXiv preprint arXiv:2110.09317},
  year={2021}
}

@article{artime2021percolation,
  title={Percolation on feature-enriched interconnected systems},
  author={Artime, Oriol and De Domenico, Manlio},
  journal={Nature communications},
  volume={12},
  number={1},
  pages={1--12},
  year={2021},
  publisher={Nature Publishing Group}
}

@article{cancho2001small,
  title={The small world of human language},
  author={Cancho, Ramon Ferrer I and Sol{\'e}, Richard V},
  journal={Proceedings of the Royal Society of London. Series B: Biological Sciences},
  volume={268},
  number={1482},
  pages={2261--2265},
  year={2001},
  publisher={The Royal Society}
}

@article{stella2018multiplex,
  title={Multiplex model of mental lexicon reveals explosive learning in humans},
  author={Stella, Massimo and Beckage, Nicole M and Brede, Markus and De Domenico, Manlio},
  journal={Scientific reports},
  volume={8},
  number={1},
  pages={1--11},
  year={2018},
  publisher={Nature Publishing Group}
}

@article{holme2005core,
  title={Core-periphery organization of complex networks},
  author={Holme, Petter},
  journal={Physical Review E},
  volume={72},
  number={4},
  pages={046111},
  year={2005},
  publisher={APS}
}

@article{veremyev2019graph,
  title={Graph-based exploration and clustering analysis of semantic spaces},
  author={Veremyev, Alexander and Semenov, Alexander and Pasiliao, Eduardo L and Boginski, Vladimir},
  journal={Applied Network Science},
  volume={4},
  number={1},
  pages={1--26},
  year={2019},
  publisher={Springer}
}

@article{comin2020complex,
  title={Complex systems: Features, similarity and connectivity},
  author={Comin, Cesar H and Peron, Thomas and Silva, Filipi N and Amancio, Diego R and Rodrigues, Francisco A and Costa, Luciano da F},
  journal={Physics Reports},
  volume={861},
  pages={1--41},
  year={2020},
  publisher={Elsevier}
}

@article{hills2009longitudinal,
  title={Longitudinal analysis of early semantic networks: Preferential attachment or preferential acquisition?},
  author={Hills, Thomas T and Maouene, Mounir and Maouene, Josita and Sheya, Adam and Smith, Linda},
  journal={Psychological science},
  volume={20},
  number={6},
  pages={729--739},
  year={2009},
  publisher={SAGE Publications Sage CA: Los Angeles, CA}
}

@article{sizemore2018knowledge,
  title={Knowledge gaps in the early growth of semantic feature networks},
  author={Sizemore, Ann E and Karuza, Elisabeth A and Giusti, Chad and Bassett, Danielle S},
  journal={Nature human behaviour},
  volume={2},
  number={9},
  pages={682--692},
  year={2018},
  publisher={Nature Publishing Group}
}

@article{boleda2020distributional,
  title={Distributional semantics and linguistic theory},
  author={Boleda, Gemma},
  journal={Annual Review of Linguistics},
  volume={6},
  pages={213--234},
  year={2020},
  publisher={Annual Reviews}
}

@article{lenci2018distributional,
  title={Distributional models of word meaning},
  author={Lenci, Alessandro},
  journal={Annual review of Linguistics},
  volume={4},
  pages={151--171},
  year={2018},
  publisher={Annual Reviews}
}

@article{beck2008mind,
  title={Mind, brain, and dualism in modern physics},
  author={Beck, Friedrich},
  journal={Psycho-Physical Dualism Today: An Interdisciplinary Approach, New York: Rowman \& Littlefield},
  pages={69--97},
  year={2008}
}

@article{steyvers2005large,
  title={The large-scale structure of semantic networks: Statistical analyses and a model of semantic growth},
  author={Steyvers, Mark and Tenenbaum, Joshua B},
  journal={Cognitive science},
  volume={29},
  number={1},
  pages={41--78},
  year={2005},
  publisher={Wiley Online Library}
}

@article{jackson2021text,
  title={From text to thought: How analyzing language can advance psychological science},
  author={Jackson, Joshua and Watts, Joseph and List, Johann-Mattis and Drabble, Ryan and Lindquist, Kristen},
  journal={Perspectives on Psychological Science},
  year={2021},
  publisher={Association for Psychological Science}
}

@article{de2008word,
  title={Word associations: Network and semantic properties},
  author={De Deyne, Simon and Storms, Gert},
  journal={Behavior research methods},
  volume={40},
  number={1},
  pages={213--231},
  year={2008},
  publisher={Springer}
}

@article{hills2021networks,
  title={Networks of the Mind: How Can Network Science Elucidate Our Understanding of Cognition?},
  author={Hills, Thomas T and Kenett, Yoed N},
  journal={Topics in Cognitive Science},
  publisher={Wiley Online Library}
}

@article{elman2004alternative,
  title={An alternative view of the mental lexicon},
  author={Elman, Jeffrey L},
  journal={Trends in cognitive sciences},
  volume={8},
  number={7},
  pages={301--306},
  year={2004},
  publisher={Elsevier}
}

@inproceedings{pennington2014glove,
  title={Glove: Global vectors for word representation},
  author={Pennington, Jeffrey and Socher, Richard and Manning, Christopher D},
  booktitle={Proceedings of the 2014 conference on empirical methods in natural language processing (EMNLP)},
  pages={1532--1543},
  year={2014}
}

@book{macwhinney2000childes,
  title={The CHILDES project: The database},
  author={MacWhinney, Brian},
  volume={2},
  year={2000},
  publisher={Psychology Press}
}

@inproceedings{whelan2015understanding,
  title={Understanding the k-medians problem},
  author={Whelan, Christopher and Harrell, Greg and Wang, Jin},
  booktitle={Proceedings of the International Conference on Scientific Computing (CSC)},
  pages={219},
  year={2015},
  organization={The Steering Committee of The World Congress in Computer Science, Computer~…}
}

@article{hadley2018sentence,
  title={Sentence diversity in early language development: Recommendations for target selection and progress monitoring},
  author={Hadley, Pamela A and McKenna, Megan M and Rispoli, Matthew},
  journal={American journal of speech-language pathology},
  volume={27},
  number={2},
  pages={553--565},
  year={2018},
  publisher={ASHA}
}

@book{fenson2007macarthur,
  title={MacArthur-Bates communicative development inventories},
  author={Fenson, Larry and others},
  year={2007},
  publisher={Paul H. Brookes Publishing Company Baltimore, MD}
}

@incollection{molloy2011critical,
  title={A critical point for random graphs with a given degree sequence},
  author={Molloy, Michael and Reed, Bruce and Newman, Mark and Barab{\'a}si, Albert-L{\'a}szl{\'o} and Watts, Duncan J},
  booktitle={The Structure and Dynamics of Networks},
  pages={240--258},
  year={2011},
  publisher={Princeton University Press}
}

@inproceedings{yamaguchi2020controlling,
  title={Controlling Internal Structure of Communities on Graph Generator},
  author={Yamaguchi, Hiroto and Ogawa, Yuya and Maekawa, Seiji and Sasaki, Yuya and Onizuka, Makoto},
  booktitle={2020 IEEE/ACM International Conference on Advances in Social Networks Analysis and Mining (ASONAM)},
  pages={937--940},
  year={2020},
  organization={IEEE}
}

@article{beckage2019network,
  title={Network growth modeling to capture individual lexical learning},
  author={Beckage, Nicole M and Colunga, Eliana},
  journal={Complexity},
  volume={2019},
  publisher={Hindawi}
}

@article{beckage2011small,
  title={Small worlds and semantic network growth in typical and late talkers},
  author={Beckage, Nicole and Smith, Linda and Hills, Thomas},
  journal={PloS one},
  volume={6},
  number={5},
  pages={e19348},
  year={2011},
  publisher={Public Library of Science San Francisco, USA}
}

@article{clerkin2017real,
  title={Real-world visual statistics and infants' first-learned object names},
  author={Clerkin, Elizabeth M and Hart, Elizabeth and Rehg, James M and Yu, Chen and Smith, Linda B},
  journal={Philosophical Transactions of the Royal Society B: Biological Sciences},
  volume={372},
  number={1711},
  pages={20160055},
  year={2017},
  publisher={The Royal Society}
}

@article{casas2018polysemy,
  title={The polysemy of the words that children learn over time},
  author={Casas, Bernardino and Catal{\`a}, Neus and Ferrer-i-Cancho, Ramon and Hern{\'a}ndez-Fern{\'a}ndez, Antoni and Baixeries, Jaume},
  journal={Interaction Studies},
  volume={19},
  number={3},
  pages={389--426},
  year={2018},
  publisher={John Benjamins}
}

@article{chang2020adjacent,
  title={Adjacent and Non-Adjacent Word Contexts Both Predict Age of Acquisition of English Words: A Distributional Corpus Analysis of Child-Directed Speech},
  author={Chang, Lucas M and De{\'a}k, Gedeon O},
  journal={Cognitive Science},
  volume={44},
  number={11},
  pages={e12899},
  year={2020},
  publisher={Wiley Online Library}
}

@inproceedings{siow2021exploring,
  title={Exploring the variable effects of frequency and semantic diversity as predictors for a word's ease of acquisition in different word classes},
  author={Siow, Serene and Plunkett, Kim},
  booktitle={Proceedings of the Annual Meeting of the Cognitive Science Society},
  volume={43},
  number={43},
  year={2021}
}

@article{pomper2019familiar,
  title={Familiar object salience affects novel word learning},
  author={Pomper, Ron and Saffran, Jenny R},
  journal={Child development},
  volume={90},
  number={2},
  pages={e246--e262},
  year={2019},
  publisher={Wiley Online Library}
}

@article{romberg2010statistical,
  title={Statistical learning and language acquisition},
  author={Romberg, Alexa R and Saffran, Jenny R},
  journal={Wiley Interdisciplinary Reviews: Cognitive Science},
  volume={1},
  number={6},
  pages={906--914},
  year={2010},
  publisher={Wiley Online Library}
}

@article{siew2021global,
  title={Global and Local Feature Distinctiveness Effects in Language Acquisition},
  author={Siew, Cynthia SQ},
  journal={Cognitive Science},
  volume={45},
  number={7},
  pages={e13008},
  year={2021},
  publisher={Wiley Online Library}
}

@article{levy2021unveiling,
  title={Unveiling the nature of interaction between semantics and phonology in lexical access based on multilayer networks},
  author={Levy, Orr and Kenett, Yoed N and Oxenberg, Orr and Castro, Nichol and De Deyne, Simon and Vitevitch, Michael S and Havlin, Shlomo},
  journal={Scientific reports},
  volume={11},
  number={1},
  pages={1--14},
  year={2021},
  publisher={Nature Publishing Group}
}

@article{zock2015words,
  title={Words in Books, Computers and the Human Mind},
  author={Zock, Michael},
  journal={Journal of Cognitive Science},
  volume={16},
  number={4},
  pages={355--378},
  year={2015}
}

@book{aitchison2012words,
  title={Words in the mind: An introduction to the mental lexicon},
  author={Aitchison, Jean},
  year={2012},
  publisher={John Wiley \& Sons}
}

@article{mcrae2005semantic,
  title={Semantic feature production norms for a large set of living and nonliving things},
  author={McRae, Ken and Cree, George S and Seidenberg, Mark S and McNorgan, Chris},
  journal={Behavior research methods},
  volume={37},
  number={4},
  pages={547--559},
  year={2005},
  publisher={Springer}
}

@book{newman2018networks,
  title={Networks},
  author={Newman, Mark},
  year={2018},
  publisher={Oxford university press}
}

@article{van2015examining,
  title={Examining assortativity in the mental lexicon: Evidence from word associations},
  author={Van Rensbergen, Bram and Storms, Gert and De Deyne, Simon},
  journal={Psychonomic bulletin \& review},
  volume={22},
  number={6},
  pages={1717--1724},
  year={2015},
  publisher={Springer}
}

@article{ferrer2018origins,
  title={The origins of Zipf's meaning-frequency law},
  author={Ferrer-i-Cancho, Ramon and Vitevitch, Michael S},
  journal={Journal of the Association for Information Science and Technology},
  volume={69},
  number={11},
  pages={1369--1379},
  year={2018},
  publisher={Wiley Online Library}
}

@article{griffiths2007google,
  title={Google and the mind: Predicting fluency with PageRank},
  author={Griffiths, Thomas L and Steyvers, Mark and Firl, Alana},
  journal={Psychological science},
  volume={18},
  number={12},
  pages={1069--1076},
  year={2007},
  publisher={SAGE Publications Sage CA: Los Angeles, CA}
}

@article{de2019small,
  title={The “Small World of Words” English word association norms for over 12,000 cue words},
  author={De Deyne, Simon and Navarro, Danielle J and Perfors, Amy and Brysbaert, Marc and Storms, Gert},
  journal={Behavior research methods},
  volume={51},
  number={3},
  pages={987--1006},
  year={2019},
  publisher={Springer}
}

@article{stella2015patterns,
  title={Patterns in the English language: phonological networks, percolation and assembly models},
  author={Stella, Massimo and Brede, Markus},
  journal={Journal of Statistical Mechanics: Theory and Experiment},
  volume={2015},
  number={5},
  pages={P05006},
  year={2015},
  publisher={IOP Publishing}
}

@book{zipf2016human,
  title={Human behavior and the principle of least effort: An introduction to human ecology},
  author={Zipf, George Kingsley},
  year={2016},
  publisher={Ravenio Books}
}

@book{miller1998wordnet,
  title={WordNet: An electronic lexical database},
  author={Miller, George A},
  year={1998},
  publisher={MIT press}
}

@article{newman2003mixing,
  title={Mixing patterns in networks},
  author={Newman, Mark EJ},
  journal={Physical review E},
  volume={67},
  number={2},
  pages={026126},
  year={2003},
  publisher={APS}
}

@article{ravasz2003hierarchical,
  title={Hierarchical organization in complex networks},
  author={Ravasz, Erzs{\'e}bet and Barab{\'a}si, Albert-L{\'a}szl{\'o}},
  journal={Physical review E},
  volume={67},
  number={2},
  pages={026112},
  year={2003},
  publisher={APS}
}

@article{bokanyi2021universal,
  title={Universal patterns of long-distance commuting and social assortativity in cities},
  author={Bok{\'a}nyi, Eszter and Juh{\'a}sz, S{\'a}ndor and Karsai, M{\'a}rton and Lengyel, Bal{\'a}zs},
  journal={Scientific reports},
  volume={11},
  number={1},
  pages={1--10},
  year={2021},
  publisher={Nature Publishing Group}
}

@article{rossetti2019cdlib,
  title={CDLIB: a python library to extract, compare and evaluate communities from complex networks},
  author={Rossetti, Giulio and Milli, Letizia and Cazabet, R{\'e}my},
  journal={Applied Network Science},
  volume={4},
  number={1},
  pages={1--26},
  year={2019},
  publisher={Springer}
}

@article{liu2009statistical,
  title={Statistical properties of Chinese semantic networks},
  author={Liu, HaiTao},
  journal={Chinese Science Bulletin},
  volume={54},
  number={16},
  pages={2781--2785},
  year={2009},
  publisher={Springer}
}

@article{mcpherson2001birds,
  title={Birds of a feather: Homophily in social networks},
  author={McPherson, Miller and Smith-Lovin, Lynn and Cook, James M},
  journal={Annual review of sociology},
  volume={},
  number={},
  pages={},
  year={2001},
  publisher={Annual Reviews 4139 El Camino Way, PO Box 10139, Palo Alto, CA 94303-0139, USA}
}

@article{bassolas2021first,
  title={First-passage times to quantify and compare structural correlations and heterogeneity in complex systems},
  author={Bassolas, Aleix and Nicosia, Vincenzo},
  journal={Communications Physics},
  volume={4},
  number={1},
  pages={1--14},
  year={2021},
  publisher={Nature Publishing Group}
}

@article{peel2018multiscale,
  title={Multiscale mixing patterns in networks},
  author={Peel, Leto and Delvenne, Jean-Charles and Lambiotte, Renaud},
  journal={Proceedings of the National Academy of Sciences},
  volume={115},
  number={16},
  pages={4057--4062},
  year={2018},
  publisher={National Acad Sciences}
}

@article{utsumi2015complex,
  title={A complex network approach to distributional semantic models},
  author={Utsumi, Akira},
  journal={PloS one},
  volume={10},
  number={8},
  pages={e0136277},
  year={2015},
  publisher={Public Library of Science San Francisco, CA USA}
}

@article{kumar2021critical,
  title={A Critical Review of Network-Based and Distributional Approaches to Semantic Memory Structure and Processes},
  author={Kumar, Abhilasha A and Steyvers, Mark and Balota, David A},
  journal={Topics in Cognitive Science},
  year={2021},
  publisher={Wiley Online Library}
}

@article{stella2019modelling,
  title={Modelling early word acquisition through multiplex lexical networks and machine learning},
  author={Stella, Massimo},
  journal={Big Data and Cognitive Computing},
  volume={3},
  number={1},
  pages={10},
  year={2019},
  publisher={Multidisciplinary Digital Publishing Institute}
}

\end{document}